%% file: main.tex
\documentclass[runningheads]{llncs}

\usepackage{eccv}

\usepackage{eccvabbrv}
\usepackage{graphicx}
\usepackage{booktabs}

\usepackage[accsupp]{axessibility}  

\usepackage{hyperref}

\usepackage{orcidlink}

\input{preamble}

\begin{document}


\title{\textsc{Appearance Pointers} \\ Multimodal Region Control of Diffusion Transformers
\vspace{-0.1in}
} 

\titlerunning{Appearance Pointers}

\author{
Rahul Sajnani$^{1,2}$\;\; Yulia Gryaditskaya$^{2}$\;\; Radomir Mech$^{2}$ \;\; \\  Srinath Sridhar$^{1}$\;\; Matheus Gadelha$^{2}$\\ \vspace{1mm}
\text{\normalsize $^1$Brown University\qquad $^2$Adobe Research}\\
\href{https://ivl.cs.brown.edu/research/appearance_pointers.html}{ivl.cs.brown.edu/research/appearance\_pointers}
\vspace{-1mm}}

\authorrunning{R. Sajnani et al.}

\institute{}

{
  \renewcommand{\addcontentsline}[3]{} 
  \maketitle
}

\addtocontents{toc}{\protect\setcounter{tocdepth}{-1}}


\begin{figure}
  \centering
  \vspace{-0.35in}
  \makebox[\textwidth][c]{\includegraphics[width=1.15\textwidth]{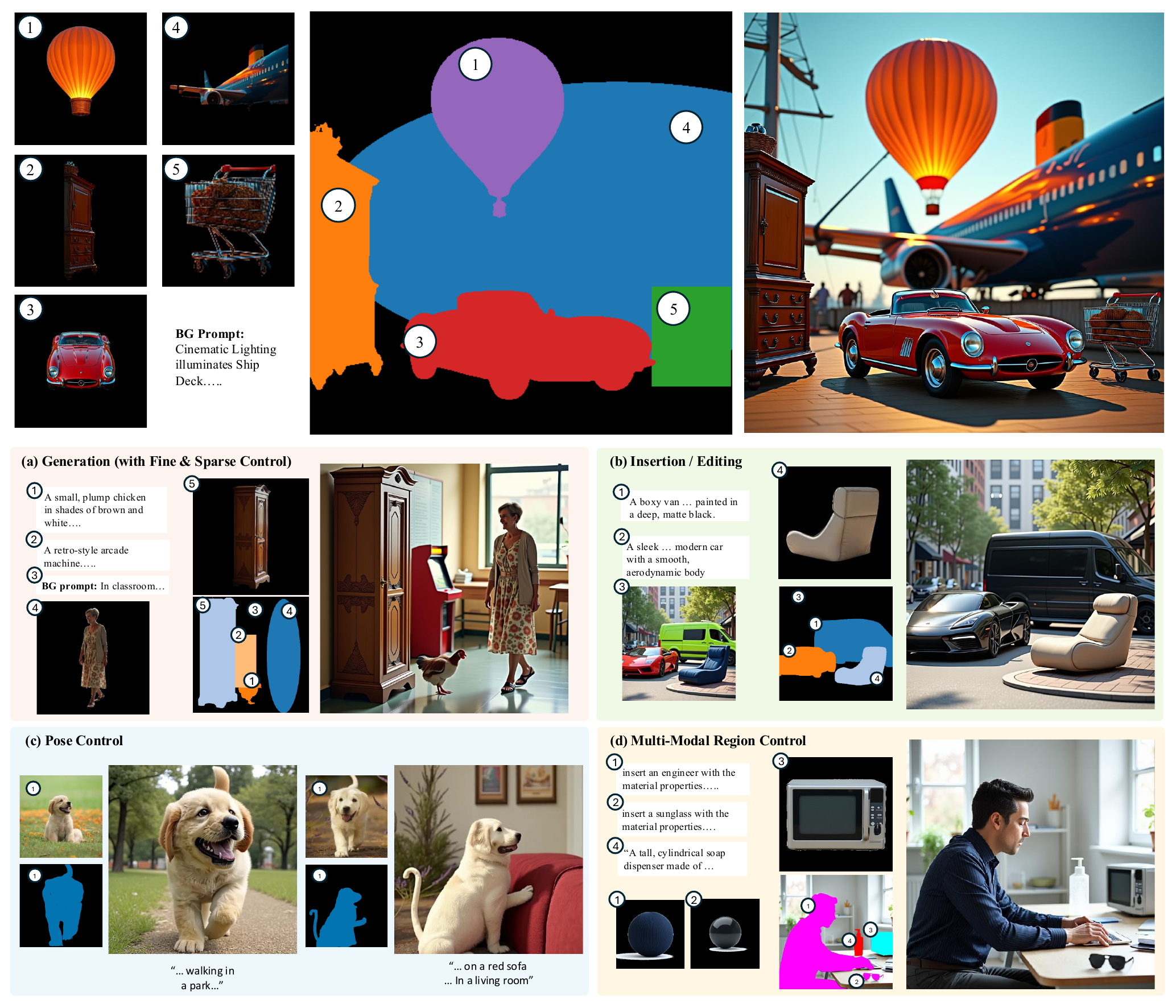}
  \vspace{-2.3in}}
  \caption{\textbf{Appearance Pointers enable precise, multimodal, region-aware image generation in a single denoising pass}.
Given spatial masks and heterogeneous conditioning signals — reference images or text descriptions — our region correspondence network produces \textbf{compact pointer tokens} that \textbf{route a Diffusion Transformer} toward the correct appearance cues at the appropriate spatial locations. 
Our work supports: \textbf{(a)} generation from fine or sparse regional layouts; \textbf{(b)} object insertion with material and style fidelity; \textbf{(c)} pose-conditioned generation; and \textbf{(d)} simultaneous multi-modal region control, combining image- and text-based descriptions in a single scene.
}
  \label{fig:teaser}
  \vspace{-0.4in}
\end{figure}

\input{sec/00_abstract}
\input{sec/01_intro}

\input{sec/02_related_works}
\input{sec/04_method}

\input{sec/05_results_and_evaluation}

\input{sec/06_conclusion}

\input{sec/07_limitations}

\clearpage
\appendix
\gdef\theHsection{\Alph{section}}
\begingroup
    \makeatletter
    \addtocontents{toc}{\protect\setcounter{tocdepth}{2}} 
    
    \renewcommand{\contentsname}{Appendix} 
    \renewcommand{\section}[2]{} 
    
    \setcounter{tocdepth}{2} 
    \tableofcontents
    \makeatother
\endgroup



\input{sec/08_supplement_mat}

%
%
\bibliographystyle{splncs04}
\bibliography{main}
\end{document}

%% file: preamble.tex
\usepackage[table]{xcolor}
\usepackage{xspace}
\usepackage{amsmath}
\usepackage{pbox}
\usepackage{algorithm}
\usepackage{algpseudocode}
\usepackage{listing}
\usepackage{listings}
\usepackage{wrapfig}
\usepackage{adjustbox}
\usepackage{booktabs}
\usepackage{makecell}
\usepackage{graphicx}
\usepackage{multirow}
\usepackage{tabularx}
\usepackage{amssymb} 
\usepackage{bm}
\newcommand{\cmark}{\textcolor{green!60!black}{\bm{$\checkmark$}}}
\newcommand{\xmark}{\textcolor{red!60!black}{\bm{$\times$}}}

\definecolor{json_key}{rgb}{0,0.44,0.34}       
\definecolor{json_string}{rgb}{0.75,0.0,0.0}   
\definecolor{json_number}{rgb}{0.0,0.0,1.0}   
\definecolor{json_background}{rgb}{0.97,0.97,0.97}

\lstdefinelanguage{json}{
    basicstyle=\normalfont\ttfamily\scriptsize,
    numberstyle=\scriptsize\color{gray},
    stepnumber=1,
    numbersep=8pt,
    showstringspaces=false,
    breaklines=true,
    backgroundcolor=\color{json_background},
    frame=single,
    rulecolor=\color{lightgray},
    string=[s]{"}{"},
    stringstyle=\color{json_key},
    morestring=[s]{""}{""},
    literate=
     *{0}{{{\color{json_number}0}}}{1}
      {1}{{{\color{json_number}1}}}{1}
      {2}{{{\color{json_number}2}}}{1}
      {3}{{{\color{json_number}3}}}{1}
      {4}{{{\color{json_number}4}}}{1}
      {5}{{{\color{json_number}5}}}{1}
      {6}{{{\color{json_number}6}}}{1}
      {7}{{{\color{json_number}7}}}{1}
      {8}{{{\color{json_number}8}}}{1}
      {9}{{{\color{json_number}9}}}{1}
      {:}{{{\color{black}{:}}}}{1}
      {,}{{{\color{black}{,}}}}{1}
      {\{}{{{\color{black}{\{}}}}{1}
      {\}}{{{\color{black}{\}}}}}{1}
      {[}{{{\color{black}{[}}}}{1}
      {]}{{{\color{black}{]}}}}{1},
}

\definecolor{goldcell}{RGB}{255, 245, 200}
\definecolor{silvercell}{RGB}{235, 235, 235}

\definecolor{cb-black}      {RGB}{  0,   0,   0}
\definecolor{cb-blue-green} {RGB}{  0,  073,  073}
\definecolor{cb-green-sea}  {RGB}{  0, 146, 146}
\definecolor{cb-rose}       {RGB}{255, 109, 182}
\definecolor{cb-salmon-pink}{RGB}{255, 182, 119}
\definecolor{cb-purple}     {RGB}{ 194,   106, 119}
\definecolor{cb-blue}       {RGB}{ 0, 109, 219}
\definecolor{cb-lilac}      {RGB}{182, 109, 255}
\definecolor{cb-blue-sky}   {RGB}{109, 182, 255}
\definecolor{cb-blue-light} {RGB}{182, 219, 255}
\definecolor{cb-burgundy}   {RGB}{146,   0,   0}
\definecolor{cb-brown}      {RGB}{146,  73,   0}
\definecolor{cb-clay}       {RGB}{219, 209,   0}
\definecolor{cb-green-lime} {RGB}{ 36, 255,  36}
\definecolor{cb-yellow}     {RGB}{255, 255, 109}
\definecolor{cb-red}{RGB}{240, 94, 0} 
\definecolor{cb-edit}     {RGB}{ 60, 170, 200}

\definecolor{material-edit}{RGB}{255, 0, 255}
\definecolor{image-edit}{RGB}{0, 255, 255}
\definecolor{prompt-edit}{RGB}{255,0,0}

\newcommand{\coolname}[0]{\emph{AppearancePointers}\xspace}

\newcommand{\parahead}[1]{\noindent\textbf{#1}:\ }

\renewcommand{\paragraph}[1]{\vspace{.25em}\noindent\textbf{#1}.}



%% file: sec/00_abstract.tex
\begin{abstract}
Controllable image generation remains challenging for creative professionals, who often require precise regional control over materials, object identities, and spatial arrangements that cannot be reliably achieved through text prompting alone.
Diffusion Transformers (DiTs) can natively ingest heterogeneous tokens stemming from texts and images, but they lack mechanisms for determining where and how these tokens should influence the output.
We introduce \emph{appearance pointers}, compact tokens that guide DiTs toward the correct appearance cues at the correct spatial locations by aligning text or image inputs with user-specified masks.
Appearance pointers are produced by a region correspondence network and refined through a spatial aggregation mechanism, enabling the model to handle multiple regional descriptions without significantly increasing token load.
Our approach introduces the first modality-agnostic interface for localized multimodal control in a DiT without retraining the base model from scratch.
Across a range of metrics, our single model reaches or surpasses the performance of modality-specific state of the art methods, offering a simple and extensible path toward precise, region-aware, multimodal guidance in generative image synthesis.

\end{abstract}

%% file: sec/01_intro.tex
\section{Introduction}
\label{sec:intro}
Recent advances in generative modeling have dramatically expanded the creative possibilities of machine-generated imagery.
Artists, designers, and filmmakers can now synthesize realistic and diverse scenes by simply providing a text description.
Yet, despite their expressive potential, current systems remain difficult to control in practice.
Creative professionals often operate with specific visual intentions --- precise materials, object layouts, and stylistic details that define the desired outcome.
In contrast, text prompting provides only indirect and often unpredictable control, requiring extensive trial and error to achieve the intended composition.
This gap between human intention and model controllability limits the integration of generative models into real-world creative pipelines.

Diffusion Transformers (DiTs) --- now core to many state-of-the-art image generators --- offer an appealing pathway toward richer controllability because they can natively ingest heterogeneous token streams, including text embeddings and image tokens.
However, they lack a mechanism to determine where and how these modality-specific tokens should influence the generated image.
Simply providing more text or image tokens does not convey the spatial intent of the user, nor does it specify which appearance cues should be used in which regions.
As a result, structured and localized user intent, expressed through text or reference images, remains difficult to translate into region-aware control.

Existing controllable generation approaches attempt to address this through attention manipulation~\cite{zhou2025dreamrenderer, zhou20253dis, zhang2025seg2any}, specialized adapters~\cite{zhang2023adding,mou2023t2i,peng2024controlnext,zhao2024uni,mo2024freecontrol,jia2024ssmg,lukovnikov2025enabling}, noise initialization strategies~\cite{mao2023semantic, mao2023guided, sun2024spatial}, or gradient-based guidance during inference~\cite{xie2023boxdiff,chen2024training,sun2024spatial, zhao2023loco,xiao2023r,couairon2023zero,phung2023grounded,lukovnikov2025enabling}.
The majority of existing work is developed for older U-Net-based diffusion models and adopts inference-time approaches that are often slow and non-robust. 
Moreover, most are constrained to a single conditioning modality (e.g., text-only or image-only) and do not seamlessly combine heterogeneous cues \cite{saha2025sigma, zhang2025seg2any}. 
Even with the adoption of recent DiT backbones, these limitations restrict the practical use of diffusion models in tasks requiring multimodal and spatially grounded control.

\input{sec/figures_and_tables_tex/fig_attention_correspondence}
Consider the image to the right where reference images were passed to a DiT to generate the image in the left column.
Yellow/green overlays on the reference images correspond to tokens being attended by probe tokens in the generated image --
represented by yellow/green stars.
Top row showcases our method while bottom row shows a vanilla baseline that simply passes reference image tokens to
the DiT during generation.
Our method attends to the corresponding regions thanks to a new type of token -- \coolname{}.
They are compact tokens designed to be ingested directly by a DiT to route the model toward the correct appearance cues at the correct spatial locations provided by the user .
Appearance pointers are produced by a small region correspondence network that fuses text or reference images with their associated spatial masks (see small color-coded regions in the top left corner of the figure above).
Rather than storing appearance directly, appearance pointers tell the DiT \emph{where} to use the vanilla image and text tokens provided by the user.
This mechanism preserves the architectural flexibility of DiTs while adding a lightweight and modular interface for localized multimodal guidance.
We also carefully designed a spatial aggregation mechanism that fuses appearance information from multiple regions,
allowing the DiT to handle several regional descriptions -- from images or text -- together,
\emph{within a single denoising process}.
Despite its simplicity, the modular design of \coolname{} allows it to be applied effectively in a variety of
workflows -- see Fig.~\ref{fig:teaser} for illustrations and Table~\ref{tab:comparison_capabilities} for a comparison
of the model capabilities with state-of-the-art approaches.
More crucially, it extends capabilities of the model to condition a specific region with both \textit{image and text simultaneously} enabling new applications such as material and text conditioned generation in ~\cref{fig:teaser} (d).


To train and evaluate this model, we created a synthetic dataset with image and text description for multiple elements
within the same image.
When regions are described only through text, our approach is the best or second best in all
the six metrics.
When images are used for regional description, our approach surpasses MSDiffusion~\cite{wang2025msdiffusion} \& DreamRenderer~\cite{zhou2025dreamrenderer} in region adherence and identity preservation.
Beyond these quantitative gains, appearance pointers offer a simple and extensible path toward precise, multimodal, region-aware control in generative modeling, enabling users to specify not only what should appear in an image, but where and how they should be realized. To sum up, our main contributions are:
\begin{itemize}
    \item We introduce \coolname, a compact representation that routes diffusion transformers to their appropriate regional signals in the form of images, text, and masks.
    \item Our modular and flexible approach extends to multiple edit capabilities including: (a) region controlled generation, (b) insertion, (c) pose control, and (d) multi-modal region-controlled generation for multiple regions in a single denoising process.
    \item We also propose a dataset \coolname-37K containing region text descriptions, appearance images from novel views, and an automatic generation scheme to generate more data.
\end{itemize}

%% file: sec/figures_and_tables_tex/fig_attention_correspondence.tex
\begin{wrapfigure}{r}{0.5\textwidth}  
  \centering
  \vspace{-0.3in}
  \includegraphics[width=\linewidth]{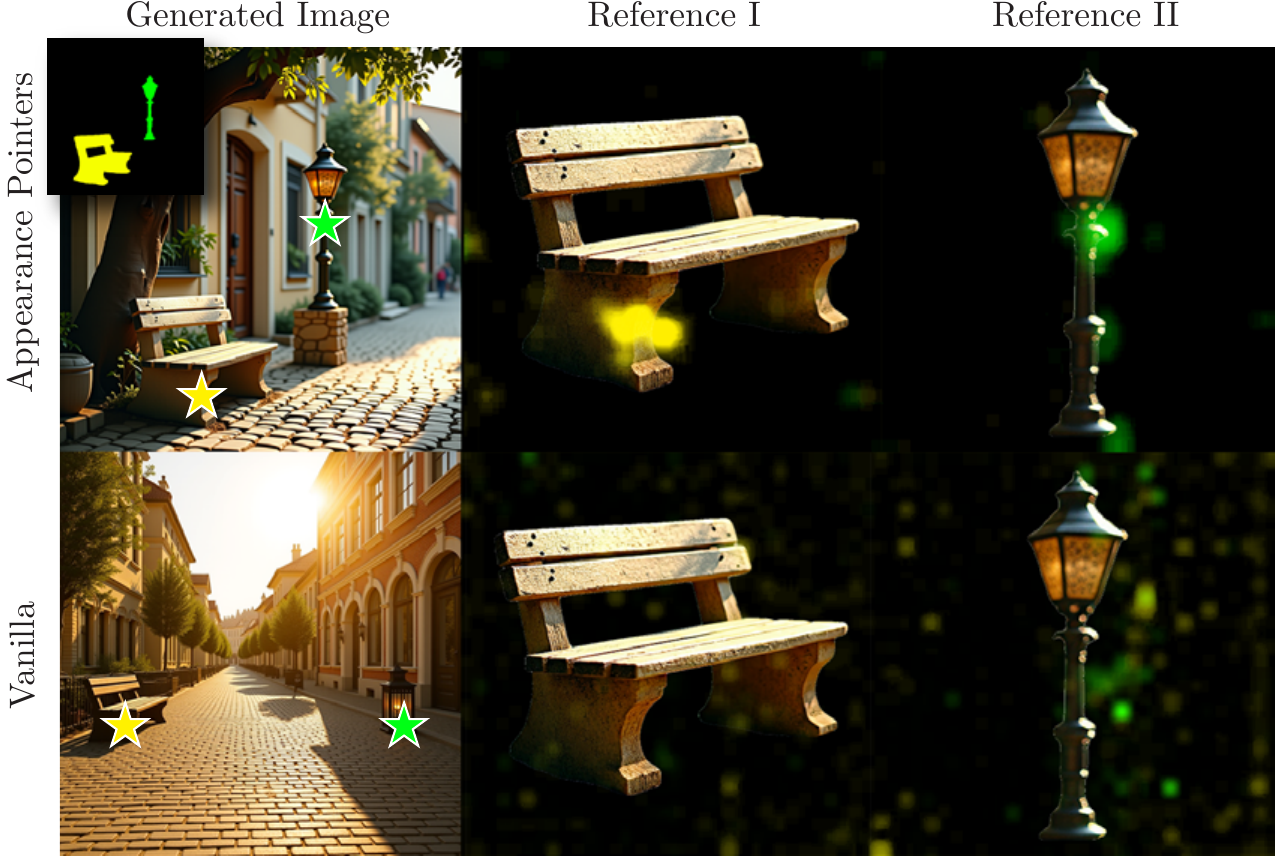}
  \vspace{-0.3in}
  \label{fig:appearance_pointer_attention_correspondence}
\end{wrapfigure}

%% file: sec/02_related_works.tex
\section{Related Works}
\label{sec:related_works}

\input{sec/figures_and_tables_tex/tab_prior_works_conditions}

For detailed discussions on multimodal controllable diffusion models,
we refer the reader to recent surveys
\cite{cao2024controllable, jiang2024survey, shuai2024survey, zhan2024conditional, wang2025image, yang2025text}.
Earlier methods operated on segmentation maps,
employing specialized architectures \cite{niu2024multi,wang2022semantic,ham2023modulating}
or general conditioning frameworks
\cite{zhang2023adding,mou2023t2i,peng2024controlnext,zhao2024uni,mo2024freecontrol,lukovnikov2025enabling},
with some extending to bounding box layouts \cite{li2021image,yang2023law}.
There are several relevant capabilities on localized text/image conditioned image generation.
One of our main goals while developing \coolname{} is to enable a variety of workflows with
the same framework without resorting to a big and complicated architecture. 
A summary of the features supported by \coolname{} and a comparison with state-of-the-art methods
can be found in Table~\ref{tab:comparison_capabilities}.
Below we discuss previous literature on region-based controls for image synthesis.

\paragraph{Bounding boxes}
Early works explored bounding-box-guided text-to-image generation
using object names \cite{mao2023training,xie2023boxdiff}
or descriptive prompts \cite{shirakawa2024noisecollage,jia2024ssmg},
often with a global scene-level prompt
\cite{li2023gligen,shirakawa2024noisecollage,xie2023boxdiff,xiao2023r,zhao2023loco,jia2024ssmg},
though these lack precise shape controllability.
\emph{Training-free} methods incorporate spatial or semantic guidance at inference time
via noise initialization \cite{mao2023semantic, mao2023guided, sun2024spatial},
cross-attention manipulation
\cite{mao2023training,chen2024training,taghipour2025box,balaji2022ediff,zhou2025dreamrenderer},
and sometimes self-attention \cite{liu2024training,zhou2025dreamrenderer,zhou20253dis}.
Taghipour et al.~\cite{taghipour2025box} use KL divergence
for stronger attribute--region association.
An alternative strategy estimates object-specific noises independently
and fuses them via cross-attention \cite{shirakawa2024noisecollage}.
Several methods apply gradient-based guidance w.r.t.\ cross-attention alignment
\cite{xie2023boxdiff,chen2024training,sun2024spatial},
cross- and self-attention consistency \cite{zhao2023loco},
or alignment augmented with a boundary-aware loss \cite{xiao2023r}.
\emph{Training-based} approaches introduce lightweight modules to fine-tune the backbone.
GLIGEN~\cite{li2023gligen} injects gated self-attention layers conditioned on grounding tokens.
SSMG~\cite{jia2024ssmg} uses a ControlNet-like branch encoding spatial and semantic features,
with attention modeling inter-object relations.
LayoutDiffusion~\cite{zheng2023layoutdiffusion} proposes a layout encoding and fusion module
with cross-attention alignment in an end-to-end diffusion framework.
Several works also target auto-regressive generation \cite{cui2025layoutenc,zheng2025layout}.
While our method supports both coarse regions (like bounding boxes) and also
has the capability of following precise regions. Additionally, it can be employed
in both image generation and editing workflows.

\paragraph{Semantic layouts}
Many training-free methods manipulate cross-attention
to strengthen concept influence within regions \cite{he2023localized,kim2023dense},
optionally restricting self-attention across segments \cite{kim2023dense}.
Gradient-based guidance updates latents via text--latent cross-attention alignment
\cite{couairon2023zero,phung2023grounded,lukovnikov2025enabling}
and sometimes self-attention \cite{phung2023grounded},
image-space losses projected to latent space \cite{bansal2023universal},
or a lightweight aligner module \cite{liu2023late}.
One approach \cite{qi2023layered} generates objects independently in early denoising steps
before merging, while MultiDiffusion \cite{bar2023multidiffusion}
optimizes the image at each noise step for global consistency.
Training-based methods integrate attention manipulation to fine-tune the base model
\cite{xue2023freestyle, wang2023enhancing, zhang2025eligen},
or inject text- and mask-conditioned features via convolutional layers \cite{zeng2023scenecomposer}.
Most closely related, Seg2Any \cite{zhang2025seg2any} builds on Flux
with attention restriction during training and region contour maps for boundary precision,
but does not support image references.
In contrast, \coolname{} unifies text and image regional conditioning within the same framework,
allowing heterogeneous appearance cues to be combined across regions in a single denoising pass. 

\paragraph{Reference image guidance}
For a comprehensive overview of appearance transfer, we refer to \cite{wei2025personalized}.
DreamRenderer \cite{zhou2025dreamrenderer} focuses on text-based region control
and encodes reference images into text embeddings via Redux \cite{blackforestlabs2024flux1}.
AnyDoor \cite{chen2024anydoor} uses DINOv2 \cite{Oquab2023dinov2} features injected via cross-attention
alongside ControlNet \cite{zhang2023adding}-style conditioning for masks and structure.
MimicBrush \cite{chen2024zero} employs dual U-Nets,
injecting reference attention keys and values into an imitative U-Net.
Insert Anything \cite{song2025insert} builds on Flux
with reference image, target region, and text conditioning;
Ace++ \cite{mao2025ace} uses channel-wise concatenation
but is outperformed by \cite{song2025insert}.
Unlike our work, these approaches support only \emph{a single region at a time},
whereas \coolname{} generates multiple regions (described by images or text)
in a single diffusion pass.
Furthermore, while these methods entangle conditioning signals globally or process regions sequentially,
\coolname{} routes each appearance cue to its correct spatial location via appearance pointer tokens,
enabling precise and modular multimodal control without architectural redundancy. 

%% file: sec/figures_and_tables_tex/tab_prior_works_conditions.tex


\begin{table}[t]
\centering
\captionsetup{width=\textwidth}
\renewcommand{\arraystretch}{1.2} 
\setlength{\tabcolsep}{3pt}      
\begin{adjustbox}{width=0.9\columnwidth}
\bfseries
\begin{tabular}{l c c c c c c c}
\toprule
\multirow{2}{*}{Method} & \multirow{2}{*}{Image} & \multirow{2}{*}{Text} & \multicolumn{2}{c}{Region Control} & \multirow{2}{*}{Insertion} & \multirow{2}{*}{Generation} & \multirow{2}{*}{MultiModal} \\
\cmidrule(lr){4-5}
& & & Fine & Sparse & & & \\
\midrule
InsertAnything~\cite{song2025insert} & \cmark & \xmark & \cmark & \cmark & \cmark & \xmark & \xmark \\
MS-Diffusion~\cite{wang2025msdiffusion} & \cmark & \xmark & \xmark & \cmark & \xmark & \cmark & \xmark \\
Sigma-Gen~\cite{saha2025sigma} & \cmark & \xmark & \cmark & \cmark & \xmark & \cmark & \xmark \\
DreamRenderer~\cite{zhou2025dreamrenderer} & \cmark & \cmark & \xmark & \cmark & \xmark & \cmark & \xmark \\
InstanceDiffusion~\cite{wang2024instancediffusion} & \xmark & \cmark & \cmark & \cmark & \xmark & \cmark & \xmark \\
Seg2Any~\cite{zhang2025seg2any} & \xmark & \cmark & \cmark & \xmark & \xmark & \cmark & \xmark \\
\rowcolor{gray!20} 
\textbf{Ours} & \cmark & \cmark & \cmark & \cmark & \cmark & \cmark & \cmark \\
\bottomrule
\end{tabular}
\end{adjustbox}
\vspace{5pt}
\caption{
\textbf{Comparison of region-conditioned generation methods across key capabilities}.
\emph{Image} and \emph{Text} indicate the supported modalities for regional appearance description.
\emph{Insertion} denotes the ability to place an object into an existing scene,
\emph{Generation} denotes full image synthesis from regional descriptions,
and \emph{MultiModal} indicates whether a method can use both image and text description \emph{at the same time for the same region} 
within a single generation pass.
This enables novel use cases like the material assignment example in Fig.~\ref{fig:arch_precisegen}.
Unlike all prior works, \coolname{} supports every capability simultaneously,
enabling flexible and spatially grounded multimodal control in a unified framework.
}
\label{tab:comparison_capabilities}
\vspace{-0.3in}
\end{table}

%% file: sec/04_method.tex
\section{Method}
\label{sec:method}
\paragraph{Preliminaries} 
\label{sec:background}
Recent state-of-the-art image generative models~\cite{blackforestlabs2024flux1, Labs2025FLUX1KF, Esser2024Scaling}
are predominantly based on transformer architectures that jointly process textual and visual information through full self-attention.
These models operate on two input modalities: a \textit{text stream}, which encodes the prompt description, and a \textit{visual stream}, which encodes conditioning images and generative tokens.
A common architectural design involves the use of a \textit{dual-transformer} scheme, in which text and image tokens are first processed by modality-specific projection layers before being fused through full self-attention.
The dual-transformer blocks promote alignment between textual and visual representations, after which a series of \textit{single-transformer} blocks perform unified multimodal processing.

\input{sec/figures_and_tables_tex/arch_global}

\paragraph{Overview}
Our method generates an image 
$I \in \mathbb{R}^{H \times W \times 3}$ that satisfies both global and regional conditions. 
The global condition is provided as a text prompt, while the regional conditions are defined as a set of $n$ pairs 
$\mathcal{R} = \{(R_i, {P}_i)\}_{i=1}^{n}$. 
Here, each $R_i \in \{0, 1\}^{H \times W}$ denotes a region mask, and $P_i$ represents the corresponding local prompt, which may consist of tokens coming from text descriptions,  reference images, or both.
The model is tasked to generate an image where each region corresponding to $R_i$ follows the content $P_i$, while the rest of the image aligns with the global text prompt (see ~\Cref{fig:arch_precisegen}).

While vanilla DiTs possess the ability of ingesting image or text tokens,
simply passing additional tokenized information is not enough to inform the models about
the localization and composition of the elements in the final image.
To this end, we introduce a new type of input token responsible for informing the DiT
where and how localized text and image information should be used during generation.
We call these \textbf{\emph{appearance pointers}}.
Their role is to ensure correct linking between regions ($R_i$) and prompts ($P_i$).
In practice, as shown in \cref{fig:arch_precisegen}, the \emph{appearance pointers} serve as a conditional input to the  DiT model, alongside global and region-specific information ---
\emph{local image prompts} and \emph{local text prompts} complement our \emph{appearance pointers} to provide fine-grained appearance details, while \emph{region contour maps} 
promote fine-grained alignment with region boundaries. 

\subsection{Appearance Pointers}
\label{sec:apperance_pointer_network}
We first describe how we construct \emph{appearance pointers}.
To align with our objectives, a good regional representation $\mathcal{R}$ should enable multi-modal control while explicitly linking the local prompts $P_i$ to the intended region $R_i$.  
To achieve this, we design a \textit{Region Correspondence Transformer} that links and processes all modalities jointly, before passing them to the DiT generative model. 
We first describe how each region $R_i$ and its corresponding prompt $P_i$ are encoded, and then explain how these representations are aligned to form the \emph{appearance pointers}. 


\paragraph{Region and Condition Encoding} 
\label{sec:encoding}
Each regional information $(R_i, P_i)$ consists of a binary mask $R_i$ and a local prompt $P_i$, which may be text, an image, or both.
Image prompts are encoded with the VAE image encoder providing image tokens $^{I}\mathcal{P}_i$, while text prompts are encoded with the Flux Kontext T5 encoder~\cite{Labs2025FLUX1KF} providing text tokens $^{T}\mathcal{P}_i$.
To encode the mask itself, we treat it as an image input to the same VAE encoder.
Before encoding, we augment the binary mask with coordinate information to provide explicit spatial grounding.
Please see supplement for more details.
The coordinate-enhanced mask $M_i$ is then encoded and patchified in the same manner as an image prompt, yielding mask tokens with explicit spatial context.

\input{sec/figures_and_tables_tex/region_prompt_figure}

\paragraph{Region-Prompt Linking} 
\label{sec:linking}
After encoding masks, images, and text to the DiT latent space, our next goal is to \textit{link} masks to their respective condition tokens. 
This \textit{linking} enables DiTs to \textit{point} to the desired condition signal and helps generate region conditioned images (see \cref{fig:appearance_pointer_attention_correspondence} for pointer attention visualization). 
To integrate spatial constraints with semantic content, we introduce a \emph{Region Correspondence Transformer}, $\mathbf{\Phi}_{RC}$ to process tokens from the mask $^{R}\mathcal{P}_i$, image $^{I}\mathcal{P}_i$, and text $^{T}\mathcal{P}_i$ prompts to produce two semantic feature maps, ${}^{I}M_i$ and ${}^{T}M_i$ for each region $i$:
\begin{align}
    {}^{I}M_i, \; {}^{T}M_i := \mathbf{\Phi}_{RC}([\, \hat{^{R}\mathcal{P}_i}, \; ^{I}\mathcal{P}_i, \; ^{T}\mathcal{P}_i \,]).
    \label{eq:correspondence_tokens}
\end{align}
These maps serve as modular semantic region representations that guide region-conditioned image generation.  
Specifically, ${}^{I}M_i$ and ${}^{T}M_i$ respectively are targeting the image and text streams of the backbone DiT model, enabling full utilization of its multi-modal conditioning mechanism.

Within $\mathbf{\Phi}_{RC}$, the region tokens undergo initial alignment via two lightweight mask transformer (2 layer self-attention) blocks aligning mask with the corresponding image and text condition.
This mask transformer also undergoes token downsampling analogous to U-DiT~\cite{Tian2024UDiTs} for efficiency (see~\cref{fig:arch_linking}).
The processed mask tokens, image, and text tokens are concatenated along the sequence dimension and provided to a multi-modal Correspondence Transformer with self-attention blocks (consistent with the Flux MM-DiT attention). 
These attention blocks use separate query, key, and value projections for each tokens type. 
In practice, we also pass learnable tokens to account for missing modalities allowing the module to operate flexibly on image+mask, text+mask, or image+text+mask inputs. 
Crucially, $\Phi_{RC}$ operates on each region independently and is diffusion-step independent. 
This enables us to run region correspondence only once per generation, significantly reducing overhead.
Additionally, this design allows the number of regions to vary dynamically during inference.

\input{sec/figures_and_tables_tex/fig_appearance_pointer_generation}

\paragraph{Region Aggregation}
\label{sec:apperance_pointers_generation}
With the regions linked to their corresponding conditions we can condition the diffusion transformer to generate region controlled images. 
However, naively injecting all region tokens along with their conditional tokens increases the token count drastically and is unfeasible with $\mathcal{O}(N^2)$ complexity for self-attention. 
Hence, we create \emph{Appearance Pointers} by consolidating information across regions using a \emph{Region Aggregation Transformer} into a single canvas producing Appearance Pointers (see \cref{fig:arch_apperance_pointers}).

To preserve  spatial structure, we perform this aggregation locally at each patch location.
We first collect each semantic region embedding ${}^{T}M_i, {}^{I}M_i \in \mathbb{R}^{N \times C}$ from \cref{eq:correspondence_tokens} and stack them along the region dimension to obtain ${}^I\mathcal{M}, {}^{T}\mathcal{M} \in \mathbb{R}^{N \times R \times C}$.
Here, $N$ is the number of spatial tokens, $R$ is the number of regions and $C$ is the channel dimension.
Next, the region aggregation transformer performs region-wise self-attention for each patch independently, treating $R$ as the sequence length.
To consolidate the information, we prepend a learnable \texttt{[CLS]} token ($\mathcal{A}$) to the region sequence at each patch. 
This "depth-wise" processing collapses the multi-region stack into a single, unified semantic canvas that matches the spatial dimensions of the backbone DiT.

In practice, we employ two independent aggregation blocks, $\Phi_A^T$ and $\Phi_A^I$, to process the text and image streams separately. 
We extract the updated \texttt{[CLS]} tokens to serve as the final appearance pointers:
\begin{align}
    {}^{T}\mathcal{AP} &:= \Phi^T_{A}([ {}^{T}A, {}^{T}\mathcal{M}]), \\
    {}^{I}\mathcal{AP} &:= \Phi^{I}_{A}([{}^{I}A, {}^{I}\mathcal{M}]).
    \label{eq:AP-main}
\end{align}
The appearance pointer aggregation module improves identity and coherence resulting better identity preservation and image coherence (Ablation \cref{tab:ablation_metrics}).

\subsection{Base DiT Conditioning}

\paragraph{Dual Stream Conditioning}
We pass the \emph{text} (${}^{T}\mathcal{AP}$)  and \emph{image} (${}^{I}\mathcal{AP}$) \emph{appearance pointers} to the image and text streams of the Flux Kontext model, respectively, to guide the generation process.  
Additionally, we include the individual local image ${}^{I}\mathcal{P}_i$ and text ${}^{T}\mathcal{P}_i$ prompt tokens.  
Concatenating all tokens yields the conditions for the image and text streams:
\begin{align}
    X_t &:= [x_t; \, {}^{I}\mathcal{AP}; \, {}^{I}\mathcal{P}_1; \dots; {}^{I}\mathcal{P}_n] \quad \text{(Image Stream)}\\
    c &:= [{}^{G}\mathcal{P}; \, {}^{T}\mathcal{AP}; \, {}^{T}\mathcal{P}_1; \dots; {}^{T}\mathcal{P}_n] \quad \text{(Text Stream)}
\end{align}
where $x_t$ are the noisy tokens.  
The denoising step of the model is then written as:
\begin{align}
    x_{t-1} := \text{FLUX}(X_t, c, t).
\end{align}
Note that some of the local prompts ${}^{I}\mathcal{P}_i$ and ${}^{T}\mathcal{P}_i$ may be empty.

\paragraph{Region Contour Guidance}
\label{sec:region_contour_guidance}
The \emph{appearance pointers} effectively link image regions with image and text prompts. 
However, its token-reduction design of the mask transformer and region aggregation occasionally smooth over fine-grained details.
To address this limitation, inspired by DreamRenderer~\cite{zhou2025dreamrenderer} and Seg2Any~\cite{zhang2025seg2any}, we aggregate all edges into a single boundary map, which is encoded using the FLUX VAE and provided to the DiT for improved precision (Ablation~\cref{tab:ablation_metrics}).

\paragraph{Implementation Details}
We train \coolname using a flow-matching objective with log-normal time sampling on 8$\times$A100 GPUs for three days.  
Following OminiControl~\cite{tan2025ominicontrol}, we apply LoRA with rank 128 to the newly introduced conditional \emph{Appearance Pointer} tokens and \emph{edge} tokens.  
All learnable parameters are trained using Prodigy~\cite{Mishchenko2023ProdigyAE} with a learning rate of $1.0$.  
Additional training details and pseudo-code are provided in the supplement.


\paragraph{Complexity} 
In terms of memory, our Region Aggregation and Region Correspondence modules consist of $\sim$400M parameters -- a 3.33\% increase over the number of parameters of the base model. 
In terms of time, it is important to highlight that Appearance Pointers are computed once for the whole inference procedure rather than per denoising timestep. 
Moreover, the design we propose with spatial downsampling and independent region processing reduces the attention computation complexity from $\mathcal{O}(T\cdot(RN_{\text{reg}})^2)$ to $\mathcal{O}(R(N_{\text{reg}}/k)^2)$ , where $k$ is the downsampling factor, $R$ is the number of regions, $N_{\text{reg}}$ is the number of tokens per region, ant $T$ is the number or regions. 
This ensures negligible overhead relative to the base generation process maintaining high inference efficiency.

\section{\textit{\coolname-37K} Data Generation}
\label{sec:dataset_generation}

\input{sec/figures_and_tables_tex/fig_dataset_example}

Existing personalization and object-insertion datasets~\cite{tan2025ominicontrol, saha2025sigma, zhang2025seg2any} lack fine-grained region captions.
We instead generate a fine-grained, region-level editing dataset enabling controlled variations in object orientation, texture, and material, with coverage across small, medium, and large objects.
We generate the dataset in three stages, described below and visualized in \cref{fig:dataset_precisegen}.

\paragraph{Scene Captioning}
We begin by categorizing everyday objects into three scale groups: small (graspable objects), medium (indoor items such as beds, tables, and chairs), and large (vehicles and infrastructure such as cars, buses, cranes, and buildings).
For each scene, we sample a variable number of objects from these categories and provide a coarse scene context description (\eg, ``kitchen'', ``living room'') to an LLM.
Using Qwen 3 \cite{yang2025qwen3}, we generate:
(i) a global scene description, and
(ii) object-level descriptions that include visual attributes, a material description, and a dedicated description for accurate object grounding.



\paragraph{Generation and Grounding}
The structured scene prompt is provided to Flux.1 Dev \cite{blackforestlabs2024flux1} to generate an image.
We then segment each described object using Grounded SAM \cite{ren2024grounded}, guided by dedicated descriptors.
This yields region masks linking each generated region to its text reference.


\paragraph{Editing and VLM Checking}
Following SIGMA-Gen \cite{saha2025sigma}, we extract each grounded object and generate two types of controlled edits:
(i) pose perturbations, and
(ii) material and texture variations, produced using Flux Kontext.
Each edited crop is evaluated by an InternVL \cite{chen2024internvl} vision–language model, verifying whether the edits remain consistent with the original descriptions through VLM-based scoring.
Full pipeline details and prompting templates are provided in the supplementary material.

%% file: sec/figures_and_tables_tex/arch_global.tex
\begin{figure*}[t]
    \vspace{-0.1in}
  \centering
  \includegraphics[width=\textwidth]{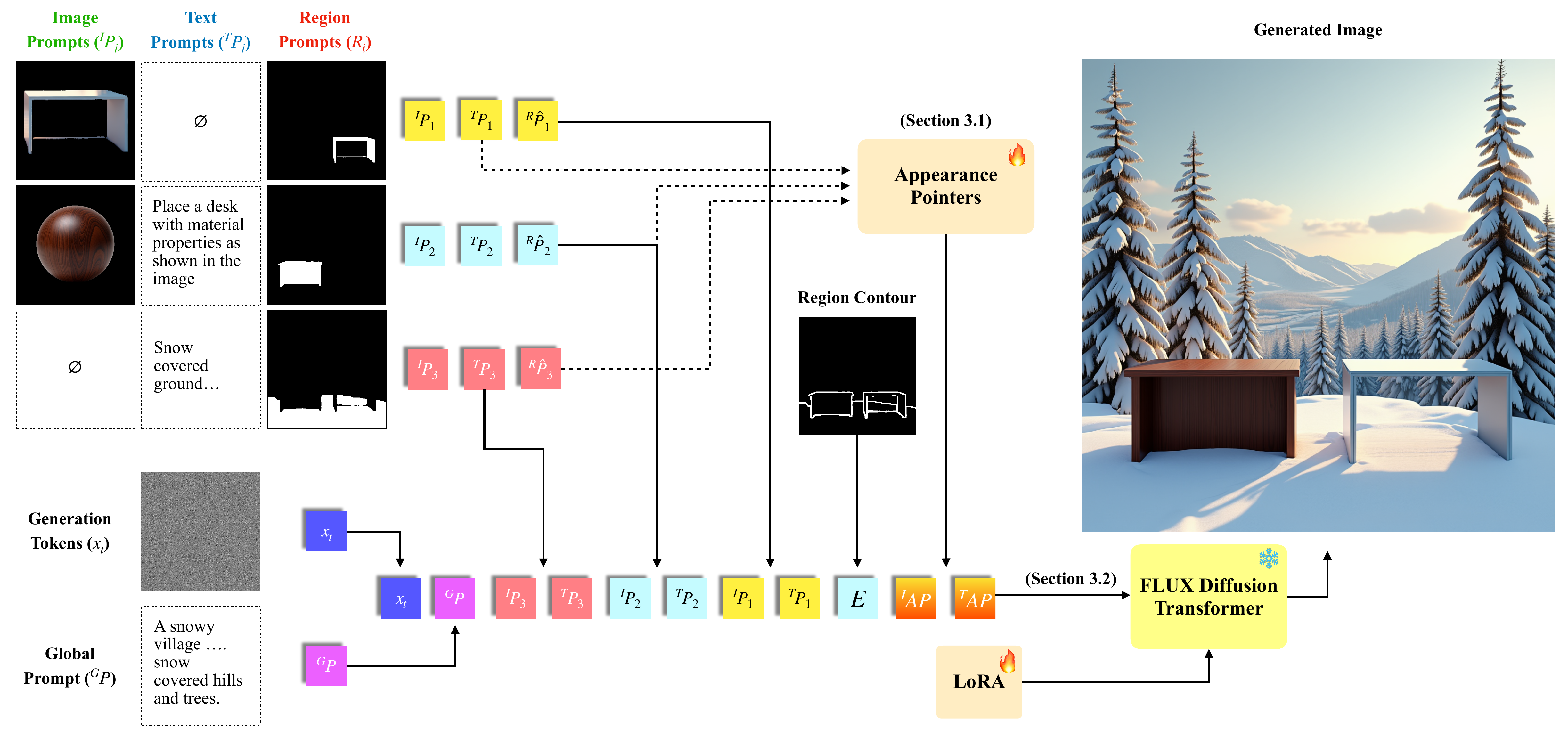}
  \vspace{-0.2in}
  \caption{
    \textbf{Overview of \emph{\coolname}}. Our framework converts heterogeneous inputs (text, images, and spatial masks) into \emph{appearance pointers} (yellow). 
    After encoding via FLUX VAE and T5 text encoder, multimodal signals from \textcolor{cb-green-sea}{image (${}^{I}P_i$)} and \textcolor{cb-blue}{text (${}^{T}P_i$)}, are fused with their \textcolor{cb-red}{spatial masks ($R_i$)} to form \textcolor{orange}{Appearance Pointers (${}^{I}\mathcal{AP}$, ${}^{T}\mathcal{AP}$)}, dictating exactly \emph{which} features map to specific regions. 
    During sampling, the diffusion transformer receives noisy tokens, text/image tokens, and the appearance pointers as joint inputs to then synthesize a coherent and spatially aligned output that respects all region-specific instructions across modalities.
    See \cref{sec:method} for more details.
  }
  \vspace{-0.2in}
  \label{fig:arch_precisegen}
\end{figure*}

%% file: sec/figures_and_tables_tex/region_prompt_figure.tex
\begin{figure*}[t]
  \centering
  \includegraphics[width=\textwidth]{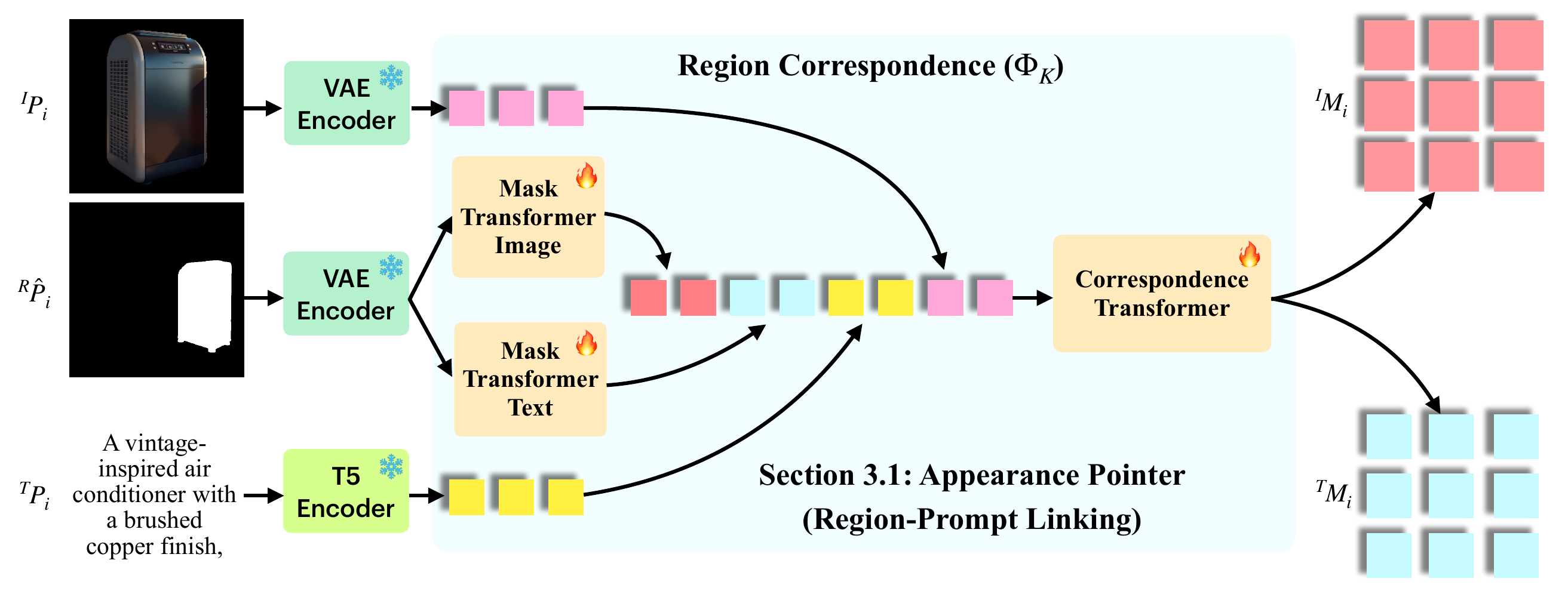}
  \vspace{-0.8cm}
  \caption{
  \textbf{Region-Prompt Linking.}
  To enable the full utilization of the multi-modal conditioning mechanism of the DiT generative backbone, we generate a set of feature maps targeting the image and text streams of the backbone DiT model. 
  Note that either the image or the text prompts can be omitted, but we will always generate feature maps targeting the two conditioning streams of the DiT model. 
  Refer to \cref{sec:encoding} for details. 
  }
  \label{fig:arch_linking}
  \vspace{-0.2in}
\end{figure*}

%% file: sec/figures_and_tables_tex/fig_appearance_pointer_generation.tex
\begin{wrapfigure}{r}{0.65\linewidth}
  \centering
  \vspace{-0.4in}
  \includegraphics[width=\linewidth]{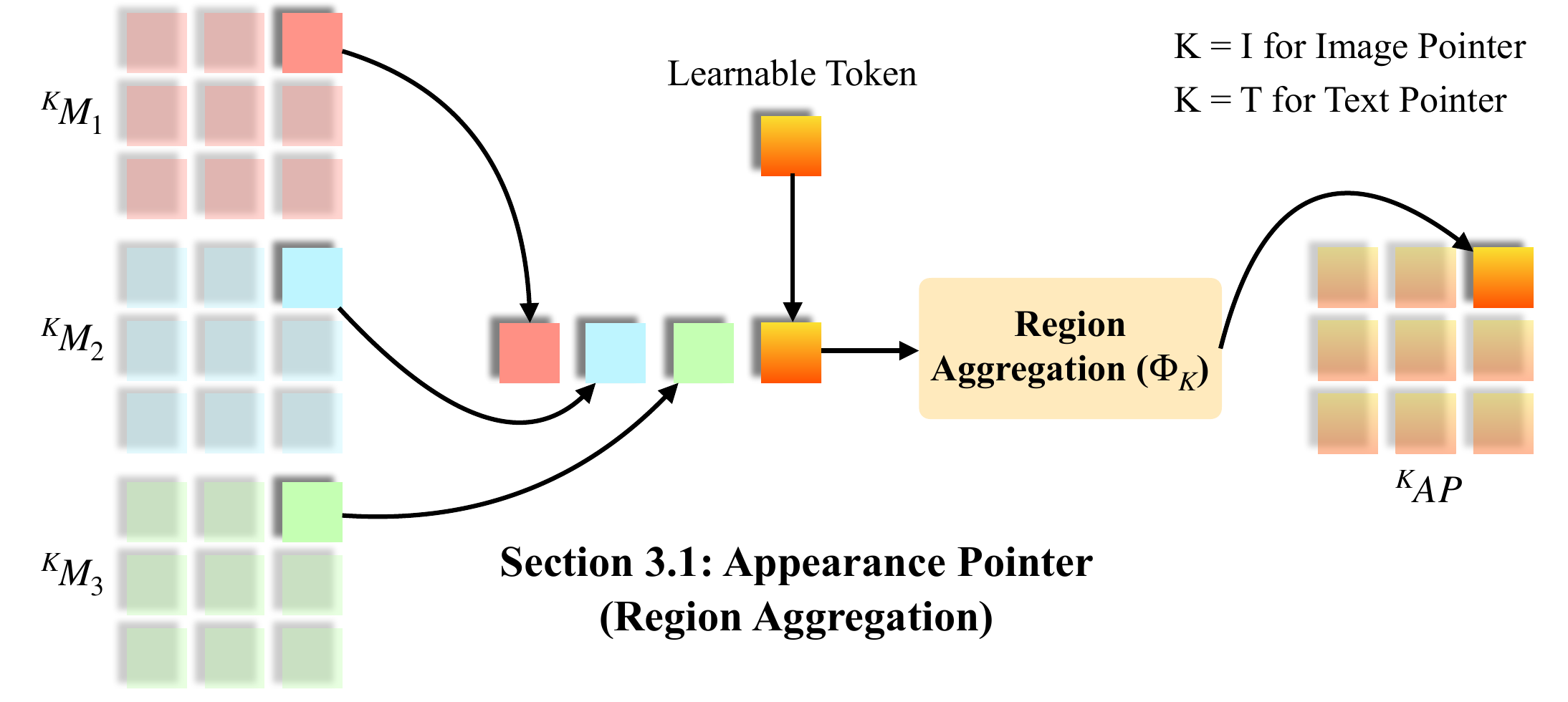}
  \vspace{-0.2in}
  \caption{
  \textbf{Appearance pointers generation.}
  The \emph{Region Aggregation Transformer} creates a concise representation of regional conditions, \emph{Appearance Pointers}.
    }
  \label{fig:arch_apperance_pointers}
  \vspace{-0.25in}
\end{wrapfigure}

%% file: sec/figures_and_tables_tex/fig_dataset_example.tex
\begin{wrapfigure}{r}{0.6\textwidth}  
    \vspace{-20pt}
  \centering
  \vspace{-0.1in}
  \includegraphics[width=\linewidth]{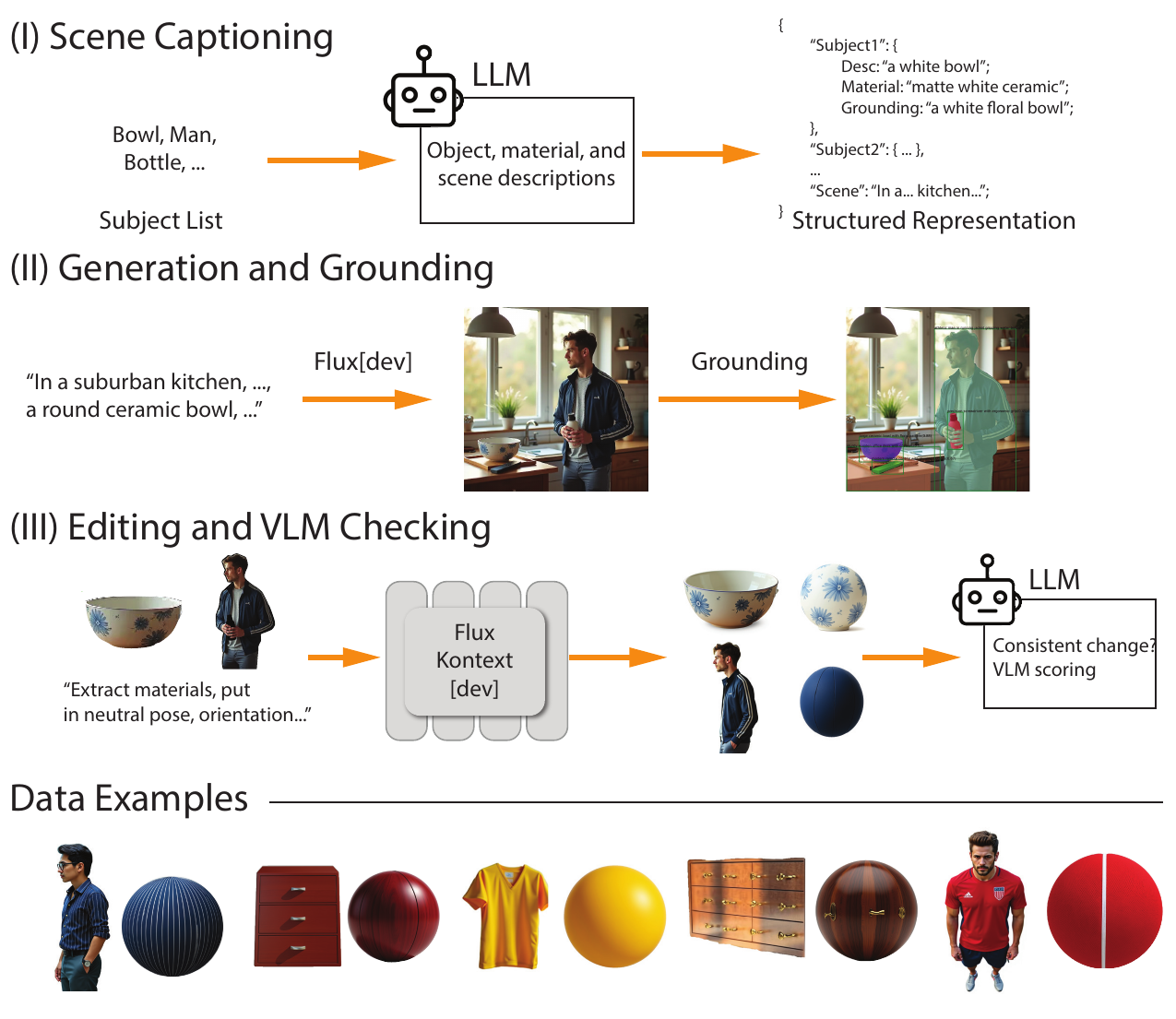}
  \vspace{-0.8cm}
  \caption{
  \textbf{Overview of the dataset creation pipeline.}
%
(I) We generate diverse scenes by sampling objects of varying scales and prompting an LLM to produce scene- and object-level descriptions.
(II) Images are synthesized using Flux.1 Dev and grounded with Grounded SAM.
(III) For each object, we produce pose and material edits and filter them using a VLM.
  }
  \label{fig:dataset_precisegen}
\vspace{-30pt}
\end{wrapfigure}

%% file: sec/05_results_and_evaluation.tex
\input{sec/figures_and_tables_tex/fig_qualtitative}

\section{Experiments \& Results}

\subsection{Quantitative Results}

\paragraph{Dataset} 
\input{sec/figures_and_tables_tex/fig_qualitative_image}
We benchmark our multi-modal region correspondence model on our \textit{Appearance Pointers-37K} (\cref{sec:dataset_generation}) dataset for region, text, and image conditioned generation.


\paragraph{Baselines} 
As prior works do not support conditioning regions on both text and images, we benchmark \coolname against unimodal regional models: \emph{InstanceDiffusion} \cite{wang2024instancediffusion}, \emph{DreamRenderer} \cite{zhou2025dreamrenderer}, and \emph{Seg2Any} \cite{zhang2025seg2any}, --- for text+region-to-image generation (\cref{tab:text_metrics} and \cref{fig:qualitative_precisegen_text}).
%
To evaluate image+region-to-image generation, we also compare \coolname against the prior state-of-the-art method, \emph{MS-Diffusion} \cite{wang2025msdiffusion} and DreamRenderer*~\cite{zhou2025dreamrenderer} (\cref{tab:image_metrics} and \cref{fig:qualitative_precisegen}).  

In addition to the described generation setting here, we also evaluate an editing variant of \coolname that preserves background regions of an input image while performing targeted modifications.  
This allows the model to maintain existing content, providing finer control for region-specific edits.  
For further implementation details, additional examples and comparison against InsertAnything~\cite{song2025insert}, we refer the reader to the supplementary material.



\paragraph{Metrics} 
Our evaluation benchmark consists of generating 500 images under two conditions: text-only and image-only, both guided by global prompts with an average of 5 regions per image.  
For the image-only condition, the benchmark additionally includes novel viewpoints of the objects to be inserted into the specified regions.

First, we evaluate our results using two \emph{global metrics}: 
(i) \textbf{CLIP-IQA}~\cite{wang2023exploring}, which assesses visual quality by comparing the CLIP embedding of the generated image with a `quality' prompt;  
(ii) \textbf{CLIP-T (Global)}, which measures alignment between the generated image and the global text prompt via the cosine similarity of their CLIP embeddings.
We further evaluate \emph{region precision} using four metrics: 
(i) \textbf{DINO-I}, which measures fidelity by computing the cosine similarity between the CLS tokens of DINO features from the generated and ground truth image patches;  
(ii) \textbf{CLIP-I}, analogous to DINO-I but using CLIP image embeddings;  
(iii) \textbf{CLIP-T}, which assesses alignment between the generated image region and its corresponding region prompt via cosine similarity of CLIP embeddings;  
(iv) \textbf{Class-agnostic MIoU}~\cite{zhang2025seg2any}, computed by prompting SAM2 on the ground truth regions within the generated image to evaluate shape consistency.

\paragraph{Analysis}
\Cref{tab:text_metrics} and \Cref{tab:image_metrics} compare our method against prior works for generation with text-conditioned regions and image-conditioned regions, respectively.
Qualitative comparisons are shown in \cref{fig:qualitative_precisegen_text} and \cref{fig:qualitative_precisegen}.
%

As shown in \Cref{tab:text_metrics}, on text-conditioned region generation, our method achieves the highest scores in global image quality (CLIP-IQA: 95.02), region fidelity (CLIP-I: 90.40), and semantic alignment (DINO-I: 56.09) for text-region-to-image synthesis.
We are also competitive in MIoU (40.35), indicating strong adherence to region masks.
In contrast, Seg2Any and DreamRenderer often fail to strictly adhere to the specified region masks, which leads to lower MIoU scores despite competitive CLIP-T performance.

For the image-region control benchmark (\Cref{tab:image_metrics}), our method outperforms both MS-Diffusion and DreamRenderer*, achieving a CLIP-I score of 93.29, MIoU of 40.97, and DINO-I score of 69.31. 
Our approach demonstrates both higher semantic alignment and improved adherence to specified region boundaries.
Please see the supplement for more details and qualitative and quantitative comparison for subject insertion. 
Please see \cref{sec:appendix_comparison} for more experiments, baseline settings, and real-world comparisons on SACap dataset.


\input{sec/figures_and_tables_tex/text_generation_table}

\input{sec/figures_and_tables_tex/table_image_metrics}

\input{sec/figures_and_tables_tex/table_ablation}
\parahead{Ablations}
\Cref{tab:ablation_metrics} presents ablations of our \coolname method for the image-conditioned prompt setting.
We use the image-conditioned prompt setting to evaluate the effectiveness of our method in preserving identity.
We first remove the region aggregation (\cref{sec:apperance_pointers_generation}), and instead of \emph{Appearance Pointers} ${}^{T}\mathcal{AP}$ and ${}^{I}\mathcal{AP}$, directly use ${}^{T}\mathcal{M} $ and ${}^{I}\mathcal{M}$ to condition the DiT.
Region Aggregation improves subject identity (from DINO-I 54.47 to 69.31) and image quality (93.48 to 95.57).
The Region Contour improves the MIoU from 35.81 to 40.97 improving region precision.
See supplement for more detailed ablation and analysis of our work.

%% file: sec/figures_and_tables_tex/fig_qualtitative.tex
\begin{figure*}[ht]
  \centering
  \vspace{-0.3in}
  \includegraphics[width=0.95\textwidth]{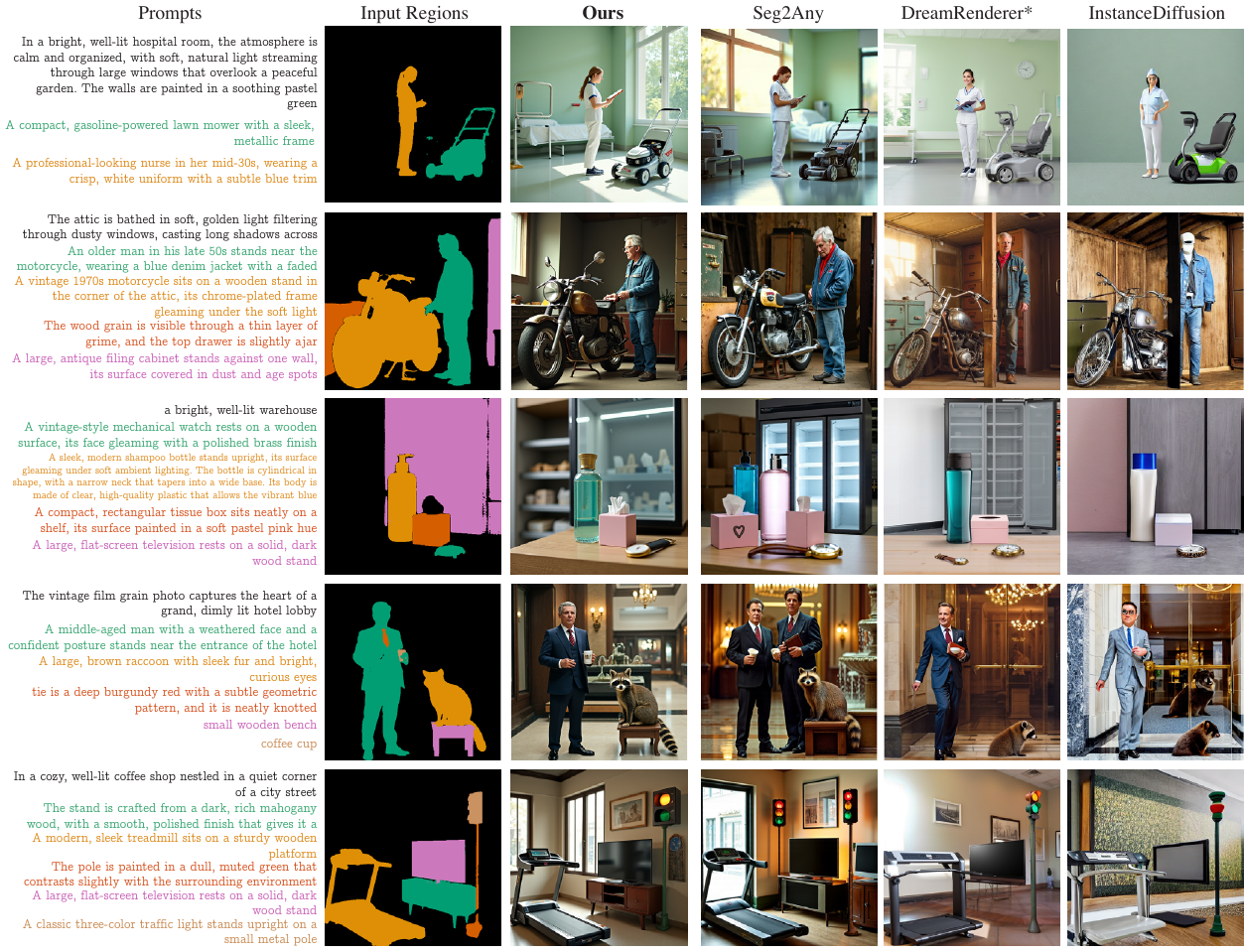}
  \vspace{-0.12in}
  \caption{
  \textbf{Qualitative comparison on region-conditioned generation from textual descriptions.}
For each example, the left column shows the user-provided regional text prompts (color-coded for clarity), followed by the corresponding input region masks. 
We compare against Seg2Any, DreamRenderer\textsuperscript{*}, and InstanceDiffusion, all of which support region-level textual guidance. 
Across diverse scenes---including indoor environments, object–person interactions, retail shelves, and multi-object arrangements---our results show stronger regional fidelity, more accurate object appearance, and better alignment with both global and local textual cues. 
DreamRenderer\textsuperscript{*} additionally uses depth information, giving it access to geometric cues not available to other baselines. 
Despite this, our model produces images with higher semantic correctness and more consistent placement across regions.
  }
  \vspace{-0.5in}
  \label{fig:qualitative_precisegen_text}
\end{figure*}

%% file: sec/figures_and_tables_tex/fig_qualitative_image.tex
%
\begin{wrapfigure}{r}{0.6\textwidth} 
  \vspace{-40pt} 
  \centering
  \includegraphics[width=\linewidth]{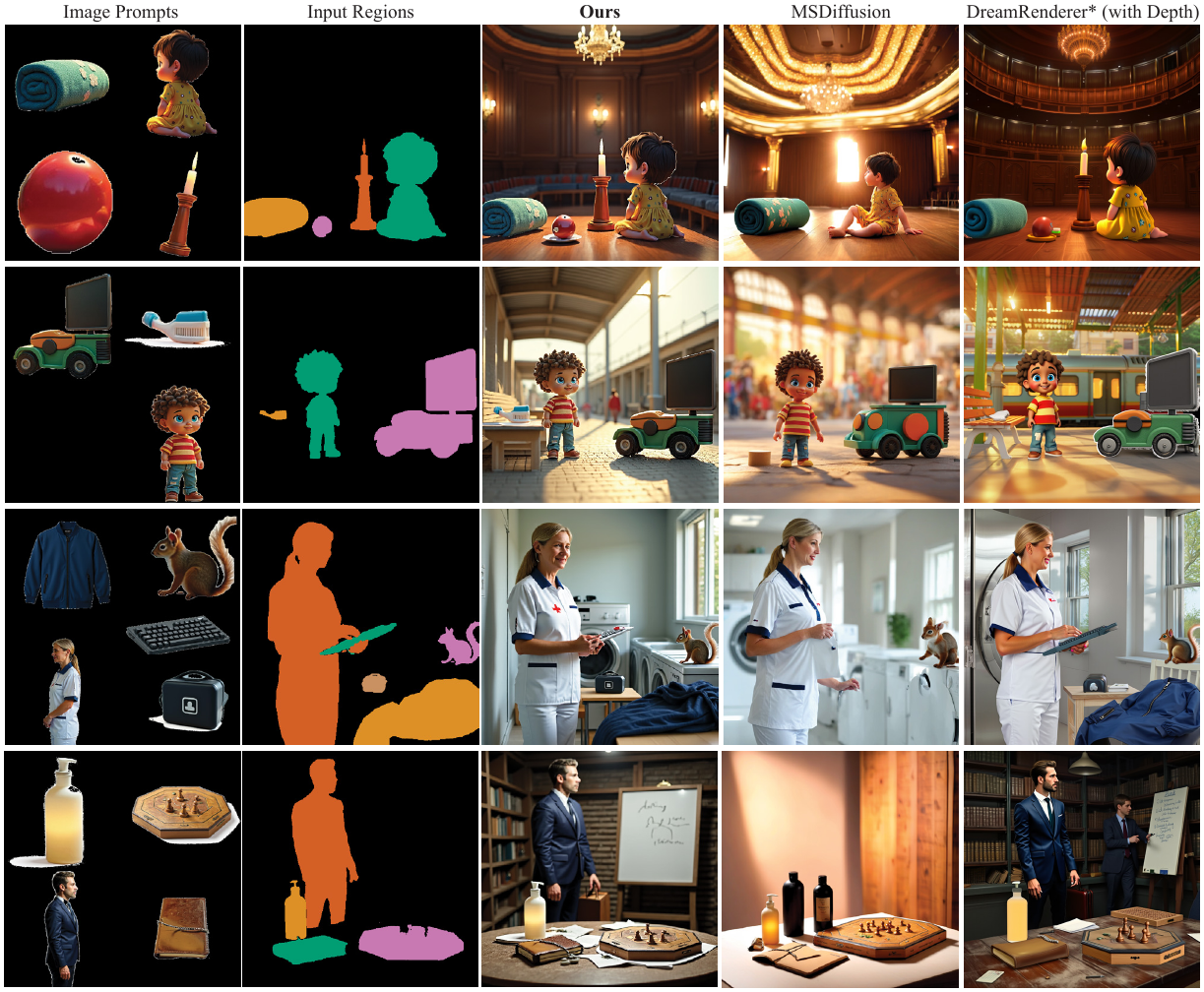}
  \vspace{-20pt} 
  \caption{\textbf{Qualitative comparison on region-conditioned generation from image prompts.}
Each example provides object reference images \textbf{(left)} and user-defined region masks \textbf{(second column)}, specifying where each object should appear in the final scene.
We compare against MSDiffusion, a recent multi-subject generation method \& DreamRenderer* that has depth supervision.
Across varied scenarios, our results show more accurate object identity and more faithful spatial placement, while MSDiffusion \& DreamRenderer often alter appearance or places objects inconsistently.
}
  \vspace{-25pt} 
   \label{fig:qualitative_precisegen}
\end{wrapfigure}

%% file: sec/figures_and_tables_tex/text_generation_table.tex
\begin{table}[t]
\centering
\vspace{-0.1in}
\caption{Quantitative comparison on AppearancePointers-37K (textual region descriptions). 
Best result in gold, second best in silver.
Higher is better for all metrics.}
\label{tab:text_metrics}
\vspace{-0.1in}
\resizebox{\linewidth}{!}
{
\begin{tabular}{lcccc|cc}
\toprule
 &
\multicolumn{4}{c}{\textbf{Region}} &
\multicolumn{2}{c}{\textbf{Global}} \\
&
\textbf{CLIP-I$\uparrow$} &
\textbf{CLIP-T$\uparrow$} & \textbf{DINO-I$\uparrow$} & \textbf{MIoU$\uparrow$} &
\textbf{CLIP-T$\uparrow$} & \textbf{CLIP-IQA$\uparrow$} \\
\midrule
InstanceDiffusion~\cite{wang2024instancediffusion} 
& 86.39 & \cellcolor{silvercell}27.54 & 35.02 & \cellcolor{goldcell}41.04 & 26.70 & 93.37 \\
DreamRenderer~\cite{zhou2025dreamrenderer} 
& 87.75 & \cellcolor{goldcell}28.03 & 44.20 & 33.84 & 28.43 & 93.24 \\
Seg2Any~\cite{zhang2025seg2any} 
& \cellcolor{silvercell}89.11 & 27.67 & \cellcolor{silvercell} 50.05 & 36.37 & \cellcolor{goldcell} 29.30 & 94.59 \\
\midrule
\textbf{Ours} 
& \cellcolor{goldcell} 90.40 & 27.24 & \cellcolor{goldcell}56.09 & \cellcolor{silvercell}40.35 & \cellcolor{silvercell} 28.93 & \cellcolor{goldcell}95.02 \\
\bottomrule
\end{tabular}
}
\vspace{-0.1in}
\end{table}

%% file: sec/figures_and_tables_tex/table_image_metrics.tex
\begin{table}[t]
\centering
\caption{Quantitative comparison on AppearancePointers-37K (image region descriptions).
\label{tab:image_metrics}
Best results in gold. }
\vspace{-0.1in}
\resizebox{\linewidth}{!}
{
\begin{tabular}{lcccc|cc}
\toprule
 &
\multicolumn{4}{c}{\textbf{Region}} &
\multicolumn{2}{c}{\textbf{Global}} \\
&
\textbf{CLIP-I$\uparrow$} &
\textbf{CLIP-T$\uparrow$} & \textbf{DINO-I$\uparrow$} & \textbf{MIoU$\uparrow$} &
\textbf{CLIP-T$\uparrow$} & \textbf{CLIP-IQA$\uparrow$} \\
\midrule
MSDiffusion~\cite{wang2025msdiffusion} & 89.66 & 24.89 & 45.61 & 28.86 &\cellcolor{goldcell} 31.08 & 74.85 \\
DreamRenderer*~\cite{zhou2025dreamrenderer} & \cellcolor{silvercell} 92.08 & \cellcolor{goldcell}27.37 & \cellcolor{silvercell} 64.20 & \cellcolor{silvercell} 40.11 & \cellcolor{silvercell} 30.02 & \cellcolor{silvercell} 89.56\\
Ours & \cellcolor{goldcell}93.29 &\cellcolor{silvercell} 27.25 & \cellcolor{goldcell}69.31 & 4\cellcolor{goldcell}0.97  &  29.62 & \cellcolor{goldcell}95.57 \\
\bottomrule
\end{tabular}
}
\vspace{-0.2in}
\end{table}

%% file: sec/figures_and_tables_tex/table_ablation.tex
\begin{table*}[t]
\centering
\vspace{-0.1in}
\caption{
\textbf{Ablation study on image generation using image-conditioned prompts.}
We evaluate the contribution of each component in the \emph{Appearance Pointer} model using the image-based regional description setting.
Removing the appearance pointer aggregation significantly harms identity preservation (lower CLIP-I and DINO-I), while slightly improving MIoU due to the absence of region fusion.
Omitting the appearance pointer mask reduces spatial disentanglement, yielding weaker region alignment.
Disabling position ID resampling degrades both identity and region consistency, especially in multi-region scenes.
Finally, removing region contour guidance leads to substantial drops in MIoU and region fidelity.
Together, these results highlight the importance of each component and confirm that the full appearance pointer formulation achieves the strongest balance of regional accuracy, spatial adherence, and identity preservation.
}
\label{tab:ablation_metrics}
\vspace{-0.1in}
\resizebox{\linewidth}{!}
{
\begin{tabular}{lcccc|cc}
\toprule
 &
\multicolumn{4}{c}{\textbf{Region}} &
\multicolumn{2}{c}{\textbf{Global}} \\
&
\textbf{CLIP-I$\uparrow$} &
\textbf{CLIP-T$\uparrow$} & \textbf{DINO-I$\uparrow$} & \textbf{MIoU$\uparrow$} &
\textbf{CLIP-T$\uparrow$} & \textbf{CLIP-IQA$\uparrow$} \\
\midrule
\textbf{w/o Region Aggregation ~\cref{sec:apperance_pointers_generation}}
& 89.93 & \cellcolor{goldcell} 27.35 & 54.47 & \cellcolor{goldcell} 41.49 & 28.45 & 93.48\\
\textbf{w/o Region Contour Guidance~\cref{sec:region_contour_guidance}} &85.57 & 23.89 & 41.62 & 35.81 & 28.89 & 94.47\\ 
\textbf{w/o Position ID Resampling Multi-Region Mimic~\cref{sec:position_id_resampling}} &  93.01 & 27.2 & 67.46 & 40.75 & 29.47 & 95.34 \\
\textbf{w/o Appearance Pointer Mask~\cref{sec:appearance_pointer_mask}} & \cellcolor{silvercell} 93.23 & \cellcolor{silvercell} 27.25 &\cellcolor{silvercell} 68.93 & \cellcolor{silvercell} 41.09 & \cellcolor{silvercell} 29.60 & \cellcolor{goldcell} 95.83 \\
\textbf{\coolname (Ours)} & \cellcolor{goldcell} 93.29 & \cellcolor{silvercell} 27.25 & \cellcolor{goldcell} 69.31 & 40.97  & \cellcolor{goldcell} 29.62 & \cellcolor{silvercell} 95.57 \\
\bottomrule
\end{tabular}
}
\vspace{-0.2in}
\end{table*}


%% file: sec/06_conclusion.tex
\section{Conclusion}
We introduced appearance pointers, a lightweight and modality-agnostic mechanism that enables Diffusion Transformers to interpret and apply region-specific user intent from text or image inputs.
By routing the model toward the correct appearance cues at the correct spatial locations, appearance pointers complement the native multimodal flexibility of DiTs.
Our region correspondence network and spatial aggregation module allow multiple regional descriptions to be combined within a single denoising process, supporting scalable and robust spatial control.
In order to train and evaluate our model, we created a synthetic dataset with text and image
descriptions of multiple regions in generated images.
When compared to modality-specific approaches, our model reaches or surpasses the state-of-the-art in most of the metrics, despite being trained for multiple modalities.
These findings demonstrate that appearance pointers provide a simple, extensible, and effective interface for precise multimodal guidance, enabling generative models to better realize not only \emph{what} users intend to create, but \emph{where} and \emph{how} the content should appear.

%% file: sec/07_limitations.tex
\paragraph{Limitations}
We observe that our work occasionally overlooks finer regions and does not preserve finer-grained details, like identity of human faces.
Moreover, we noticed degradation in the ability to follow regional information when a bigger number of regions is being prescribed; e.g. 10 or more regions.
Please see supplement for visual failure cases and its discussion.

%% file: sec/08_supplement_mat.tex
\section{Implementation Details \& Pseudocode}
\subsection{Training Setting} 
We train with up to 7 regions (including background) at each training iteration, where one of the regions can be a background region. 
During training, we randomly replace the global prompt with a background only global prompt that does not describe the foreground objects $50\%$ of the time. 
Using background prompt as the global prompt forces region modules to attend more to the region prompts and allows the model to work robustly in scenarios where no global prompt is provided.


As described in section 4 in the main manuscript, we perform VLM checking for generated (i) pose variation and (ii) material and texture variation. 
We use the generated result for training only when InternVL~\cite{chen2024internvl} consistency scores exceed $0.84$ for pose variation and $0.79$ for texture/material variation.
The equation below provides the sampling probability $p$ of a region prompt being a subject image from the same viewpoint as in the groundtruth target, a subject image from a novel viewpoint, a material reference image, a text, or their combination:
\footnotesize
\begin{equation}
^{I}\mathcal{P}, ^{T}\mathcal{P} =
\begin{cases}
p\in[0,0.4],  
    \begin{cases} 
        p\in [0.0,0.2] \text{ \& (if consistent novel view)} \\ \;\; \text{novel view image \textbf{(I + T)}}   \\
        p\in (0.2, 0.4] \text{ \& (if consistent material)}\\ \;\; \text{material image \textbf{(I + T)}}  \\
        \text{else, image crop (no text prompt) \textbf{(I)}}
    \end{cases}
    \\
p\in(0.4,0.6],  \text{image crop (no text prompt) \textbf{(I)}} \\
p \in (0.6,1.0],  \text{text only prompt \textbf{(T)}},
\end{cases}
\label{eq:sampling_prob}
\end{equation}
\normalsize
%
where \textbf{(I)} denotes \emph{image} prompt and \textbf{(T)} denotes \emph{text} prompt with cases of \textbf{(I+T)} denotes both image and text prompts. 
During training, we center each reference subject in the prompt image and apply a black background mask before passing it to our model.
Additionally, we sample coarse masks 30\% of the time where the coarse mask can be a bounding box or an ellipse.
Additionally, we provide the category name for the material prompt as the material alone does not denote the type of subject to be inserted.



\subsection{Conditions' RoPE Positional IDs}
Following~\cite{tan2025ominicontrol, saha2025sigma}, we assign distinct, non-overlapping positional embeddings to each conditioning image and to the text-prompt tokens. 
For an image prompt, the positional IDs are defined as
\[
[1,\, R(i) \cdot H + h,\, w],
\]
and for text tokens as
\[
[0,\, R(i) \cdot H + \text{mean}({}^{T}h),\, 0],
\]
where \(R(i)\) is the integer region identifier and \(H\) is the height dimension of the positional-embedding grid for the generated image, and \(h, w\) are the row and column coordinates of the token within the region. 
Here, $\text{mean}({}^{T}h)$ is the mean height of the region prompt linking tokens (${}^{T}\mathcal{M}_i$) corresponding to the regional text prompt ${}^{T}\mathcal{P}_i$.
The \emph{appearance pointer} tokens use positional IDs:
$
[1,\, h,\, w]
$
for the image appearance pointer, and
$
[0,\, h,\, w]
$
for the text appearance pointer.


\paragraph{ID Resampling}
\label{sec:position_id_resampling}
To mimic a larger number of regions in a small-region training setting, we multiply the region position ID by a random integer multiplier. 
This allows the model to experience a wider range of positional IDs, which is necessary for scaling up at inference time.  

With this approach, the image prompt token positional ID becomes
\[
[1,\, R(i) \cdot m \cdot H + h,\, w],
\]
and the regional text prompt token positional ID becomes
\[
[0,\, R(i) \cdot m \cdot H,\, 0],
\]
where $m$ is a randomly chosen integer in the range \(m \in[1,6]\).

\subsection{Appearance Pointer Mask} 
\label{sec:appearance_pointer_mask}
In the presence of multiple regions, the appearance region pointer aggregation may find it difficult to select the most relevant features. 
We assist the appearance pointer aggregation by masking regions of interest using the input region mask. 
Before aggregation all region conditions to obtain the image (${}^{I}\mathcal{AP}$) and text (${}^{T}\mathcal{AP}$) appearance pointers (defined in the main document), we re-purpose the input regional masks (${}^{R}\mathcal{P}$) to ensure non-zero features only in the prompts designated spatial regions.
This is implemented as follows:
\begin{align}
    {}^{T}\mathcal{AP} &:= \Phi^{T}_{A}([{}^{T}A, {}^{T}\mathcal{M} \odot {}^{R}\mathcal{P}]), \\
    {}^{I}\mathcal{AP} &:= \Phi^{I}_{A}([{}^{I}A, {}^{I}\mathcal{M} \odot {}^{R}\mathcal{P}]).
\end{align}
%
The region-linked features ${}^{T}\mathcal{M} \in \mathbb{R}^{R \times C \times H_t \times W_t}$ and ${}^{I}\mathcal{M} \in \mathbb{R}^{R \times C \times H_i \times W_i}$, the \emph{Region Aggregation Transformer} networks $\Phi^{I}_{A}$ and $\Phi^{T}_{A}$, and the learnable aggregation tokens ${}^{I}A$ and ${}^{T}A$ are defined in the main document.
We omit the downsample function used to resize regional masks ${}^{R}\mathcal{P}$ for clarity.
$\odot$ represents the Hadamard Product.

\subsection{Architecture}
The region correspondence block is a six layer multimodal transformer. 
Each transformer block within region correspondence has separate query, key, and value projections for text, image, and mask tokens. 
We also have a modulation block that takes the region's clip feature vector to modulate the region similar to attention in FLUX Kontext dev. 
The tokens are then passed through an MLP with skip connection.
Our hidden dimension is 768 and we perform a bottleneck operation that converts from the high dimensional FLUX embedding (D=3072) to 768. 
This makes our approach lightweight and fast without any perceivable change in performance.
The region aggregation block is constructed in a similar way with the only difference that we don't have modulation and instead skip connections with layer scale~\cite{layerscale}.
The attention of the region aggregation block is unimodal and much more lighter compared to correspondence block.

    

\input{sec/figures_and_tables_tex/fig_edit}

\section{Multi-Subject Insertion (Editing)}
A benefit of our modular region approach is that we can treat the background image and global text prompt as a region condition and seamlessly extend our approach to \emph{an editing task} that allows multi-subject insertion (see \Cref{algo:editing_line}).

We therefore benchmark our method against \emph{InsertAnything}~\cite{song2025insert}, which supports object insertion within a scene image.
Since \emph{InsertAnything} can only insert one object at a time, we construct a baseline by iteratively inserting objects. This procedure gives \emph{InsertAnything} a conceptual advantage compared to our method, which inserts all objects simultaneously.

\Cref{tab:editing_metrics} and \Cref{fig:qualitative_edit_precisegen} show quantitative and qualitative results comparing our method against \emph{InsertAnything}-based baseline. 
\coolname outperforms iterative \emph{InsertAnything} on CLIP-I, CLIP-T, MIoU, and CLIP-IQA, while remaining competitive on DINO-I, despite handling all regions at once, rather than resorting to an iterative approach. 
Moreover, the quality of the \emph{InsertAnything}'s generated images deteriorates as the number of insertion operations increases.

\input{sec/figures_and_tables_tex/table_editing}


\section{Ablations}
\Cref{tab:ablation_metrics_supp} presents ablations of our \coolname method for the image-conditioned prompt setting.
We use the image-conditioned prompt setting to evaluate the effectiveness of our method in preserving identity.


\input{sec/figures_and_tables_tex/table_ablation_metrics_supp}

\paragraph{w/o Appearance Pointer Aggregation:} 
We remove the aggregation step, and instead of \emph{Appearance Pointers} ${}^{T}\mathcal{AP}$ and ${}^{I}\mathcal{AP}$, directly use ${}^{T}\mathcal{M} $ and ${}^{I}\mathcal{M}$.
For these tokens, we then use the same LoRA fine-tuning approach as for the \emph{Appearance Pointers} tokens.
Without the aggregation step to \emph{Appearance Pointers} tokens, the identity of the objects is not well preserved, as indicated by the lower DINO-I (54.47) and CLIP-I (89.93) scores, but not performing Appearance Pointer aggregation enables finer region control with higher MIoU (41.49) as the model does not need to learn how to aggregate regions.


\paragraph{w/o Appearance Pointer Mask:} 
The masking strategy, described in \cref{sec:appearance_pointer_mask}, enhances subject identity preservation by ensuring that condition region tokens attend only to tokens within the target regions.
This results in a higher MIoU of $69.31$, compared to the baseline without the strategy, denoted in the table as \emph{w/o Appearance Pointer Mask}.

\paragraph{w/o Position ID Resampling Multi-Region Mimic:}
The position ID sampling described in \cref{sec:position_id_resampling}, improves identity for larger number of regions and region adherence without which we obtain a lower DINO-I score of 67.46 and a lower MIoU of 40.75


\paragraph{w/o Region Contour Guidance:}
Next, we perform an ablation study by removing the Region Contour Guidance.
\Cref{tab:ablation_metrics} shows that Region Contour Guidance improves all region metrics, increasing MIoU from $35.81$ to $40.97$.

\parahead{Performance vs. Number of Regions}
\cref{tab:ablation_num_regions} evaluates performance as the number of regions increases from 1 to 9 for image+region to image task. 
While identity preservation (CLIP-I) remains remarkably stable, regional DINO-I and mIoU show a gradual decline. 
This is a consequence of increased spatial occlusions and overlapping boundaries in crowded scenes.
\input{sec/figures_and_tables_tex/table_num_regions_ablation}

\parahead{Robustness to Mask Precision}
\cref{tab:ablation_coarse_regions} analyzes model performance across varying mask granularities (Fine vs. BBox vs. Ellipse). Our results demonstrate high stability in identity preservation (CLIP-I) and aesthetic quality (CLIP-IQA), even when using sparse inputs. 
The decrease in DINO-I and mIoU for coarse masks reflects the model’s flexibility in synthesizing novel subject orientations that respect the coarse spatial regions (ellipse \& bbox).

\input{sec/figures_and_tables_tex/table_coarse_regions}



\subsection{Appearance Pointer Pseudocode} Please see \Cref{algo:appearance_pointer} pseudocode for a detailed pseudocode for implementing attention pointer.
\input{sec/figures_and_tables_tex/algo_region_pointer}

\section{Qualitative Results}
\cref{fig:qualitative_gen_mix}, \cref{fig:qualitative_gen_mix_edit}, \cref{fig:qual_gallery_2}, \cref{fig:qual_gallery_1}, and \cref{fig:qual_gallery_3} show visuals of using our method with all conditions together, including material, image, and text conditions.
We also perform in-the-wild image editing (\cref{fig:qualitative_gen_in_the_wild_edit}) and layout-guided generation (\cref{fig:qualitative_gen_layout_gen}) on Google images.

\input{sec/figures_and_tables_tex/fig_supp_qual}
\input{sec/figures_and_tables_tex/fig_supp_qual_edit}

\input{sec/figures_and_tables_tex/fig_supp_gallery_2}
\input{sec/figures_and_tables_tex/fig_supp_gallery}
\input{sec/figures_and_tables_tex/fig_supp_gallery_3}

\input{sec/figures_and_tables_tex/fig_sup_in_the_wild_edit}
\input{sec/figures_and_tables_tex/fig_sup_layout_generation}

\hfill
\vfill
\clearpage

\section{Baseline Comparison Setting} 
\label{sec:appendix_comparison}
Following Seg2Any, MS-Diffusion, and InstanceDiffusion, we evaluate baselines using their original optimized weights. 
Retraining these models on our Appearance Pointers-37K dataset is avoided for the following reasons:

(1) \underline{\textit{Metric Neutrality:}} Our benchmark utilizes distribution-neutral metrics (mIoU, DINO-I, CLIP-I, CLIP-T, CLIP-IQA) to measure spatial+text adherence, identity preservation and aesthetics. 
We explicitly exclude distribution-sensitive metrics (FID/KID) to minimize inherent evaluation advantage.
We do recognize some advantage distribution shift advantage might still and ran experiments on real data below.
(2) \underline{\textit{Large-Scale Training:}} Prior works such as Seg2Any and MS-Diffusion train at large scale with millions of data samples. 
Retraining them risks creating weak baselines and is avoided by all prior works as well.
Notably, our method \textbf{utilizes the same model} for \textbf{both image- and text-based regional descriptions} — a significantly more difficult setting that is not supported by prior works. 
However, we also \textbf{show real-world} results below and \textbf{benchmark against Seg2Any \& MS-Diffusion on Real-World generation}.
\input{sec/figures_and_tables_tex/table_new_distribution}

\parahead{Real Data Quantitative Results}
We evaluate our method on the real-world SACap-eval dataset in both zero-shot and fine-tuned settings (\cref{tab:new_distribution}). 
To ensure a fair comparison with Seg2Any, which applies face blurring, we exclude human subjects from the evaluation and randomly sample 200 images for benchmarking. 
Our results demonstrate strong distribution transfer and effective performance with marginal fine-tuning data. See \cref{fig:rebuttal_eccv} (c).
\input{sec/figures_and_tables_tex/fig_rebuttal_eccv}

\section{Limitation \& Discussion with Future Works}
First, our method inherits the FLUX Kontext model's difficulty in preserving intricate human facial identities, a limitation that likely reflects the constraints of the FLUX training distribution. Empirically, we observe that the model preserves identity more reliably when faces occupy larger spatial regions rather than smaller ones.
Second, because our dataset verification relies on large-scale VLM-based filtering, residual errors in the verification process may persist and potentially degrade appearance-transfer performance.
Finally, conditioning on a large set of regions (about 10) can lead to leaking artefacts or degrade regions, as shown in \cref{fig:qualitative_failure}.

Our work \coolname introduces a method that allows \textit{Pointing} text/image/edit conditions to the desired region and shows competitive or better performance compared to unimodal prior works.
A future work for our generic design could be to extend our work to other forms of conditioning modalities, for example, audio, video, 3D, and 4D. 
Another possible direction is improving the efficiency of the model to make it real-time for editing and scene control applications.

\input{sec/figures_and_tables_tex/fig_sup_failure}

\input{sec/figures_and_tables_tex/fig_attention_correspondence_supp}
\section{Appearance Pointer Attention Map Analysis}
We provide the attention map visualization in~\Cref{fig:appearance_pointer_attention_correspondence_supp} to understand what the pointer tokens are attending to during generation.
The yellow queried points on the generated image and segments are queried to obtain the attention map reflecting the regions responsible for conditioning the segment. 
In each case, our appearance pointer highlights the region of correct region of interest. 
For example, the pattern on the chair, the right foot of the bench, the lamp, and the crane body each point to the corresponding conditioning signal.

\section{Additional Dataset Details}

\subsection{Dataset VLM Check}
Edits by FLUX Kontext may not necessarily be accurate and can have inconsistencies that can result in an erroneous signal for our model. 
Hence, we additionally use InternVL 3 VLM to rate the quality of the edits and prompt it to provide a summary of the reason for the rating (see examples in \cref{json:novel_pose} \& \cref{json:material}). 

\subsubsection{Novel View Verification}
\Cref{json:novel_view_rating} displays an example of prompting the InternVL~\cite{chen2024internvl} model to score the quality of the novel view generation of the subject of interest.
\Cref{json:novel_pose} displays the result obtained from the model to rate the overall consistency of the generated novel view.
\input{sec/figures_and_tables_tex/novel_view}

\subsubsection{Material/Texture Verification}
~\Cref{json:material_rating} displays an example of prompting the InternVL~\cite{chen2024internvl} to score the quality of the material/texture sphere generated by FLUX-Kontext for the subject of interest.
~\Cref{json:material} displays the result obtained from the InternVL model to rate the quality of the generated material with its reasoning.
\input{sec/figures_and_tables_tex/material}

\subsection{Prompting}
Here, we detail the additional details that we used to prompt LLM. 
~\Cref{json:prompt_generation} displays the JSON prompt provided to Qwen 3~\cite{yang2025qwen3} to generate object descriptions, material descriptions, Grounding DINO captions, and more.
In our work, we found succinct Grounding DINO captions necessary to improve ground score and succinct material descriptions to prompt FLUX Kontext~\cite{Labs2025FLUX1KF} to improve material/texture sphere generation.



\input{sec/figures_and_tables_tex/prompt_generation}

\input{sec/figures_and_tables_tex/novel_view_rating}

\input{sec/figures_and_tables_tex/material_rating}

%% file: sec/figures_and_tables_tex/fig_edit.tex
\begin{figure}[t]
  \centering
  \includegraphics[width=\linewidth]{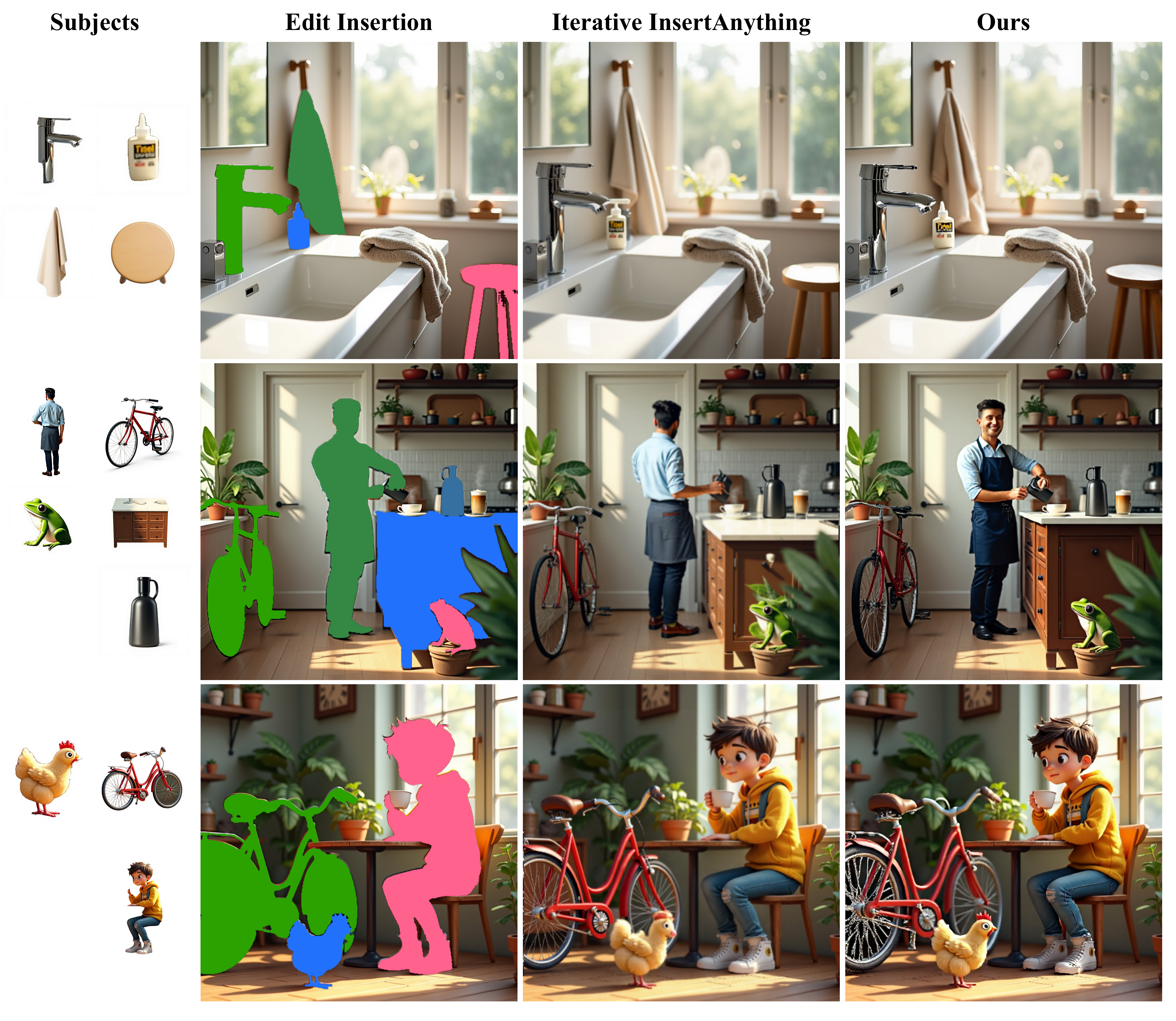}
  \caption{\textbf{Multi-Subject Insertion.} we perform multi subject insertion with \coolname and benchmark against iterative InsertAnything.
  InsertAnything tends to degrade image results after repeated subject insertion and \textbf{deforms objects such as the white bottle in row 1, the table and leaves in row 2}.
  }
  \label{fig:qualitative_edit_precisegen}
\end{figure}

%% file: sec/figures_and_tables_tex/table_editing.tex
\begin{table}[t]
\centering
\caption{Quantitative comparison on AppearancePointers-27K (Image Region Description with Background). 
Best result in gold.
Higher is better for all metrics. 
\coolname \textbf{outperforms} Iterative InsertAnything in \textbf{CLIP-I, CLIP-T, MIoU \& CLIP-IQA} and performs \textbf{competitively on DINO-I even with a setting that favors InsertAnything.}}
\label{tab:editing_metrics}
\vspace{-0.1in}
\resizebox{\linewidth}{!}
{
\begin{tabular}{lcccc|cc}
\toprule
 &
\multicolumn{4}{c}{\textbf{Region}} &
\multicolumn{2}{c}{\textbf{Global}} \\
&
\textbf{CLIP-I$\uparrow$} &
\textbf{CLIP-T$\uparrow$} & \textbf{DINO-I$\uparrow$} & \textbf{MIoU$\uparrow$} &
\textbf{CLIP-T$\uparrow$} & \textbf{CLIP-IQA$\uparrow$} \\
\midrule
\textbf{Iterative InsertAnything~\cite{song2025insert}} & \cellcolor{goldcell} 94.52 & 27.46  & \cellcolor{goldcell} 76.32 & 40.63 & 29.74 & 85.82  \\
\textbf{\coolname (Ours)} & \cellcolor{goldcell} 94.52 & \cellcolor{goldcell} 27.69 & 75.11 & \cellcolor{goldcell} 45.44 & \cellcolor{goldcell} 30.03 & \cellcolor{goldcell} 95.47\\
\bottomrule
\end{tabular}
}
\end{table}

%% file: sec/figures_and_tables_tex/table_ablation_metrics_supp.tex
\begin{table*}[t]
\centering
\caption{
\textbf{Ablation study on image generation using image-conditioned prompts.}
We evaluate the contribution of each component in the \emph{Appearance Pointer} model using the image-based regional description setting.
Removing the appearance pointer aggregation significantly harms identity preservation (lower CLIP-I and DINO-I), while slightly improving MIoU due to the absence of region fusion.
Omitting the appearance pointer mask reduces spatial disentanglement, yielding weaker region alignment.
Disabling position ID resampling degrades both identity and region consistency, especially in multi-region scenes.
Finally, removing region contour guidance leads to drop in MIoU.
Together, these results highlight the importance of each component and confirm that the full appearance pointer formulation achieves the strongest balance of regional accuracy, spatial adherence, and identity preservation.
}
\label{tab:ablation_metrics_supp}
\vspace{-0.1in}
\resizebox{\linewidth}{!}
{
\begin{tabular}{lcccc|cc}
\toprule
 &
\multicolumn{4}{c}{\textbf{Region}} &
\multicolumn{2}{c}{\textbf{Global}} \\
&
\textbf{CLIP-I$\uparrow$} &
\textbf{CLIP-T$\uparrow$} & \textbf{DINO-I$\uparrow$} & \textbf{MIoU$\uparrow$} &
\textbf{CLIP-T$\uparrow$} & \textbf{CLIP-IQA$\uparrow$} \\
\midrule
\textbf{w/o Region Aggregation~\cref{sec:apperance_pointers_generation}}
& 89.93 & \cellcolor{goldcell} 27.35 & 54.47 & \cellcolor{goldcell} 41.49 & 28.45 & 93.48\\
\textbf{w/o Appearance Pointer Mask~\cref{sec:appearance_pointer_mask}} & \cellcolor{silvercell} 93.23 & \cellcolor{goldcell} 27.25 &\cellcolor{silvercell} 68.93 & \cellcolor{silvercell} 41.09 & \cellcolor{silvercell} 29.60 & \cellcolor{goldcell} 95.83 \\
\textbf{w/o Region Contour Guidance~\cref{sec:region_contour_guidance}} &85.57 & 23.89 & 41.62 & 35.81 & 28.89 & 94.47\\ 
\textbf{w/o Position ID Resampling Multi-Region Mimic~\cref{sec:position_id_resampling}} &  93.01 & 27.2 & 67.46 & 40.75 & 29.47 & 95.34 \\
\textbf{\coolname (Ours)} & \cellcolor{goldcell} 93.29 & 27.25 & \cellcolor{goldcell} 69.31 & 40.97  & \cellcolor{goldcell} 29.62 & \cellcolor{goldcell} 95.57 \\
\bottomrule
\end{tabular}
}
\vspace{-0.3in}
\end{table*}

%% file: sec/figures_and_tables_tex/table_num_regions_ablation.tex
\begin{table}[h]
\vspace{-0.15in}
\caption{Precision vs. Number of Regions.}
\label{tab:ablation_num_regions}
\vspace{-0.15in}
\centering
\resizebox{\linewidth}{!}
{
\begin{tabular}{cllllllllll}
\toprule
\multicolumn{1}{l}{}    & Metrics / Num Regions & 1     & 2     & 3     & 4     & 5     & 6     & 7     & 8     & 9     \\ 
\midrule
\multirow{4}{*}{Region} & CLIP-T       & 30.50  & 29.81 & 29.5  & 28.96 & 28.39 & 28.19 & 26.99 & 26.41 & 25.13 \\
                        & CLIP-I $\uparrow$      & 95.07 & 95.41 & 95.41 & 95.51 & 94.87 & 95.13 & 94.84 & 95.64 & 94.94 \\
                        & DINO-I $\uparrow$  & 85.22 & 85.49 & 85.56 & 84.03 & 82.03 & 80.45 & 78.58 & 76.96 & 74.80  \\
                        & MIoU  $\uparrow$       & 51.19 & 51.95 & 51.88 & 49.13 & 48.35 & 48.05 & 46.33 & 44.13 & 43.63 \\ \midrule
\multirow{2}{*}{Global} & CLIP-T $\uparrow$      & 34.87 & 34.96 & 34.90  & 34.98 & 36.27 & 36.20  & 35.19 & 35.87 & 35.87 \\
                        & CLIP-IQA $\uparrow$     & 94.62 & 92.96 & 93.65 & 93.97 & 94.38 & 95.02 & 93.07 & 93.65 & 93.18 \\ \bottomrule
\end{tabular}
}

\vspace{-0.2in}
\end{table}

%% file: sec/figures_and_tables_tex/table_coarse_regions.tex
\begin{table}[h]
\vspace{-0.2in}
\caption{Coarse Region Control Ablation. \label{tab:ablation_coarse_regions}}
\vspace{-0.15in}
\centering
\resizebox{\linewidth}{!}
{
\begin{tabular}{lllll|ll}
\toprule
                              Ablations & \multicolumn{4}{c}{Region}       & \multicolumn{2}{c}{Global} \\ 
\midrule
Coarse/Fine Regions & CLIP-T & CLIP-I & DINO-I & MIoU  & CLIP-T      & CLIP-IQA     \\ \midrule
Blobs (Ellipse)               & 28.60   & 93.72  & 76.53  & 44.94 & 34.93       & 93.91        \\
Bounding Box (Bbox)                          & 28.79  & 94.31  & 80.32  & 47.86 & 34.91       & 93.02        \\
Fine                          & 28.93  & 95.13  & 84.03  & 53.87 & 35.00          & 93.51        \\ \bottomrule    
\end{tabular}
}
\vspace{-0.3in}
\end{table}

%% file: sec/figures_and_tables_tex/algo_region_pointer.tex
\algrenewcommand\alglinenumber[1]{\footnotesize #1:}
\begin{algorithm*}[ht]
\footnotesize
\caption{Appearance Pointer Conditioned Image Generation \label{algo:appearance_pointer}}
\begin{algorithmic}[1]
\Require ${}^{\text{R}}\mathcal{P} (\text{\footnotesize Region Masks}), {}^{\text{I}}\mathcal{P} (\text{\footnotesize Region Images}), {}^{\text{T}}\mathcal{P} (\text{\footnotesize Region Text Prompts})$ 
\Require ${}^{I}\mathcal{P}_{bg} (\text{\footnotesize Background Image}), \text{timesteps (Generation Timesteps)}$
\Require $R (\text{\footnotesize Number of Regions)}, {}^{\text{G}}\mathcal{P} (\text{\footnotesize Global Prompt})$
\Ensure $I := \text{AppearancePointerConditionedGeneration}({}^{\text{R}}\mathcal{P}, {}^{\text{I}}\mathcal{P} , {}^{\text{T}}\mathcal{P})$
\State ${}^{I}M_s := \text{List}(), \; {}^{T}M_s:=\text{List}()$
\State ${}^{R}P\text{.append}({}^{R}\mathcal{P}_{bg}), {}^{I}\mathcal{P}\text{.append}({}^{I}\mathcal{P}_{bg} \odot {}^{R}\mathcal{P}_{bg}), {}^{T}\mathcal{P}\text{.append}({}^{G}\mathcal{P})$ \Comment{Add Masked Global Image Prompt and Text Prompt for Editing Application\label{algo:editing_line}}
\For {$i=1,2,\ldots R+1$}
    \State ${}^{I}M_i, \; {}^{T}M_i := \mathbf{\Phi}_{RC}([\, {^{R}\mathcal{P}_i}, \; ^{I}\mathcal{P}_i, \; ^{T}\mathcal{P}_i \,])$ \Comment{Region-Prompt Linking where Learnable Embedding is Used if Image or Text Prompt is Missing}
    \State ${}^{I}M_s.\text{append}({}^{I}M_i), {}^{T}M_s.\text{append}({}^{T}M_i)$
\EndFor
\State ${}^{T}\mathcal{M}:=\text{Concat}({}^{T}M_s), {}^{I}\mathcal{M}:=\text{Concat}({}^{I}M_s)$
\If{\text{AppearancePointerMask}} \Comment{If Appearance-Pointer Mask}
    \State ${}^{T}\mathcal{AP} := \Phi^{T}_{A}([{}^{T}A, {}^{T}\mathcal{M} \odot {}^{R}\mathcal{P}])$
    \State ${}^{I}\mathcal{AP} := \Phi^{I}_{A}([{}^{I}A, {}^{I}\mathcal{M} \odot {}^{R}\mathcal{P} ])$
\Else
    \State ${}^{T}\mathcal{AP} := \Phi^{T}_{A}([{}^{T}A, {}^{T}\mathcal{M}])$
    \State ${}^{I}\mathcal{AP} := \Phi^{I}_{A}([{}^{I}A, {}^{I}\mathcal{M}])$
\EndIf
\State $E:= \text{VAE}({}^{R}\mathcal{P})$ \Comment{Get Contour Latent}
\State $x_{\text{timesteps}} := \mathcal{N}(0, I)$
\For {$t=\text{timesteps},\text{timesteps} - 1,\ldots 0$}
    \State $X_t := [x_t; \, {}^{I}\mathcal{AP}; \, {}^{I}\mathcal{P}_1; \dots; {}^{I}\mathcal{P}_n] \quad \text{(Image Stream)}$
    \State $c := [{}^{G}\mathcal{P}; \, {}^{T}\mathcal{AP}; \, {}^{T}\mathcal{P}_1; \dots; {}^{T}\mathcal{P}_n] \quad \text{(Text Stream)}$
    \State $x_{t-1} := \text{FLUX}(X_t, c, t, E)$ (Add Contours $E$ to Generative Model) 
\EndFor
\State I := Decode($x_0$) \Comment{Decode Generated Image}
\State \Return $I$
\end{algorithmic}
\end{algorithm*}

%% file: sec/figures_and_tables_tex/fig_supp_qual.tex
\begin{figure*}[!ht]
  \centering
  \includegraphics[width=0.95\linewidth]{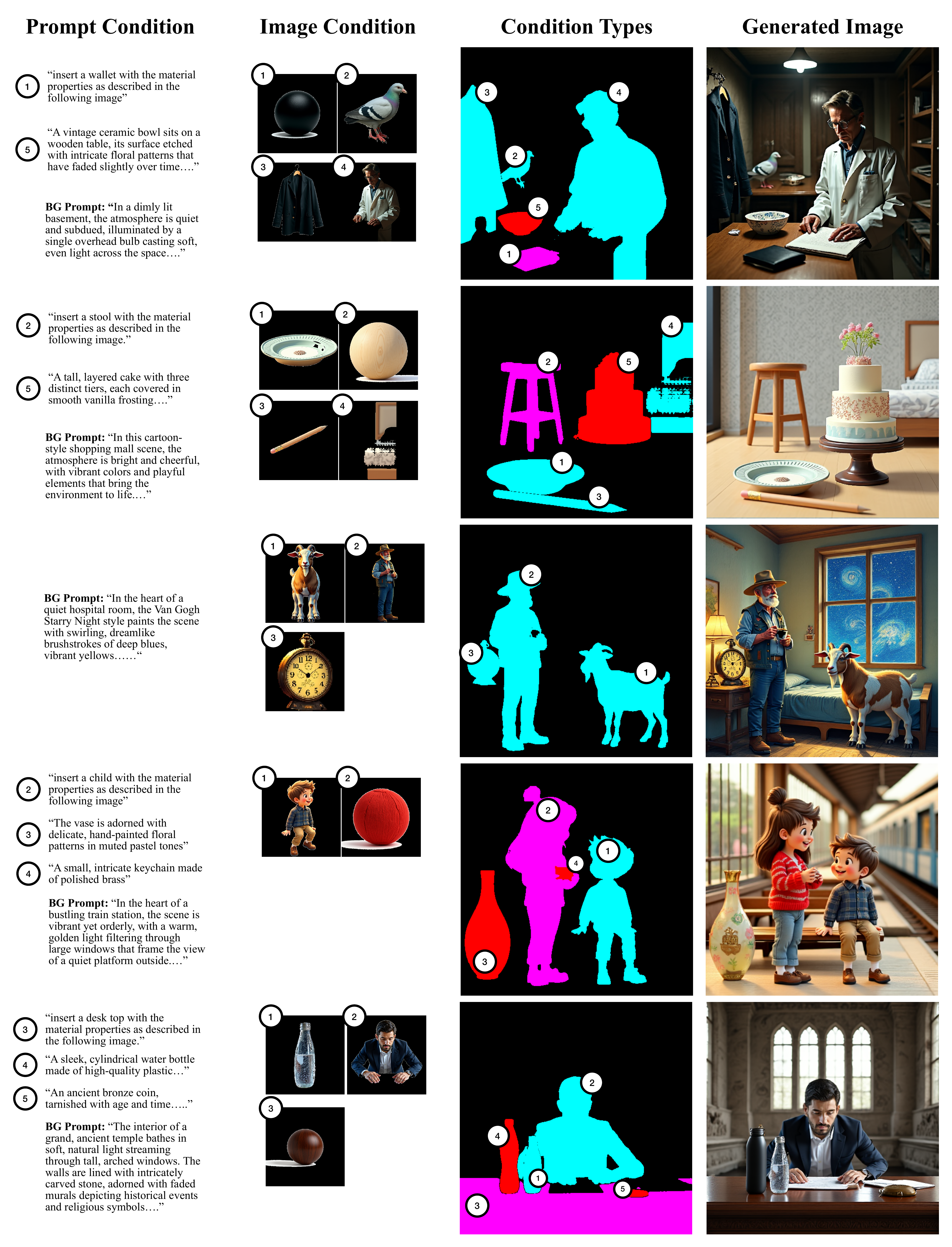}
  \vspace{-0.12in}
  \caption{\textbf{MultiModal Region Conditioned Generation.} 
  \coolname can generate images from heterogeneous condition signals that include \textcolor{material-edit}{materials}, \textcolor{image-edit}{images} and \textcolor{prompt-edit}{text prompts}. 
  Each region is numbered with the same number as the conditioning signal. The background is generated by the background description.
  Material conditioned regions are highlighted with \textcolor{material-edit}{bright purple} color, image conditioned regions are highlighted with \textcolor{image-edit}{cyan} color, and text conditioned regions are highlighted with \textcolor{prompt-edit}{red} color in column 3.
  }
  \label{fig:qualitative_gen_mix}
\end{figure*}

%% file: sec/figures_and_tables_tex/fig_supp_qual_edit.tex
\begin{figure*}[t]
  \centering
  \includegraphics[width=\linewidth]{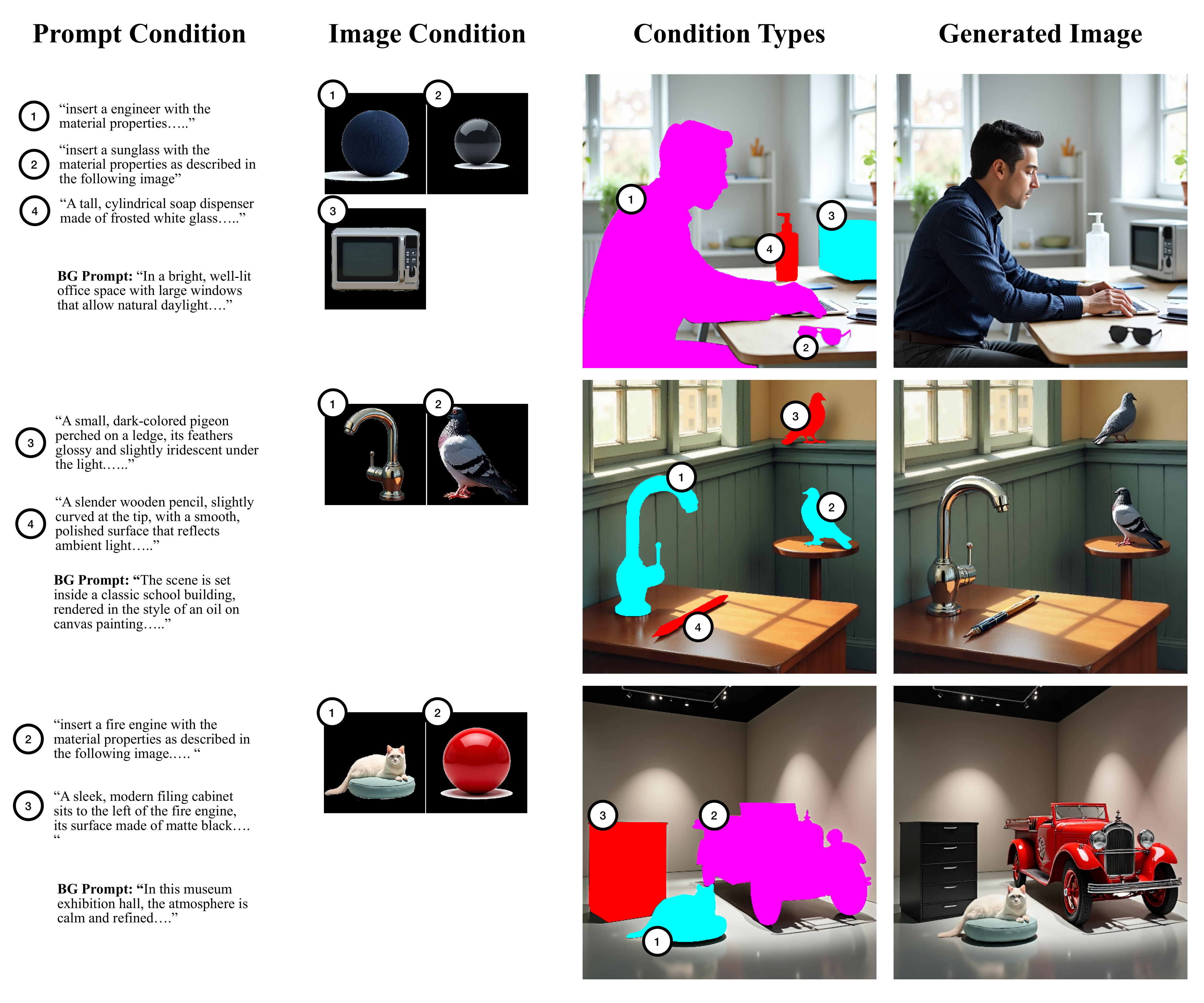}
  \vspace{-0.3in}
    \caption{\textbf{MultiModal Region Conditioned Insertion.}
  \coolname can insert subjects from heterogeneous condition signals that include \textcolor{material-edit}{materials}, \textcolor{image-edit}{images} and \textcolor{prompt-edit}{text prompts}. 
  Each region is numbered with the same number as the conditioning signal. 
  The background is unchanged for the editing scenario.
  Material conditioned regions are highlighted with \textcolor{material-edit}{bright purple} color, image conditioned regions are highlighted with \textcolor{image-edit}{cyan} color, and text conditioned regions are highlighted with \textcolor{prompt-edit}{red} color in column 3.
  }
  \label{fig:qualitative_gen_mix_edit}
\end{figure*}

%% file: sec/figures_and_tables_tex/fig_supp_gallery_2.tex
\begin{figure*}[t]
  \centering
  \vspace{-0.15in}
  \includegraphics[width=0.95\linewidth]{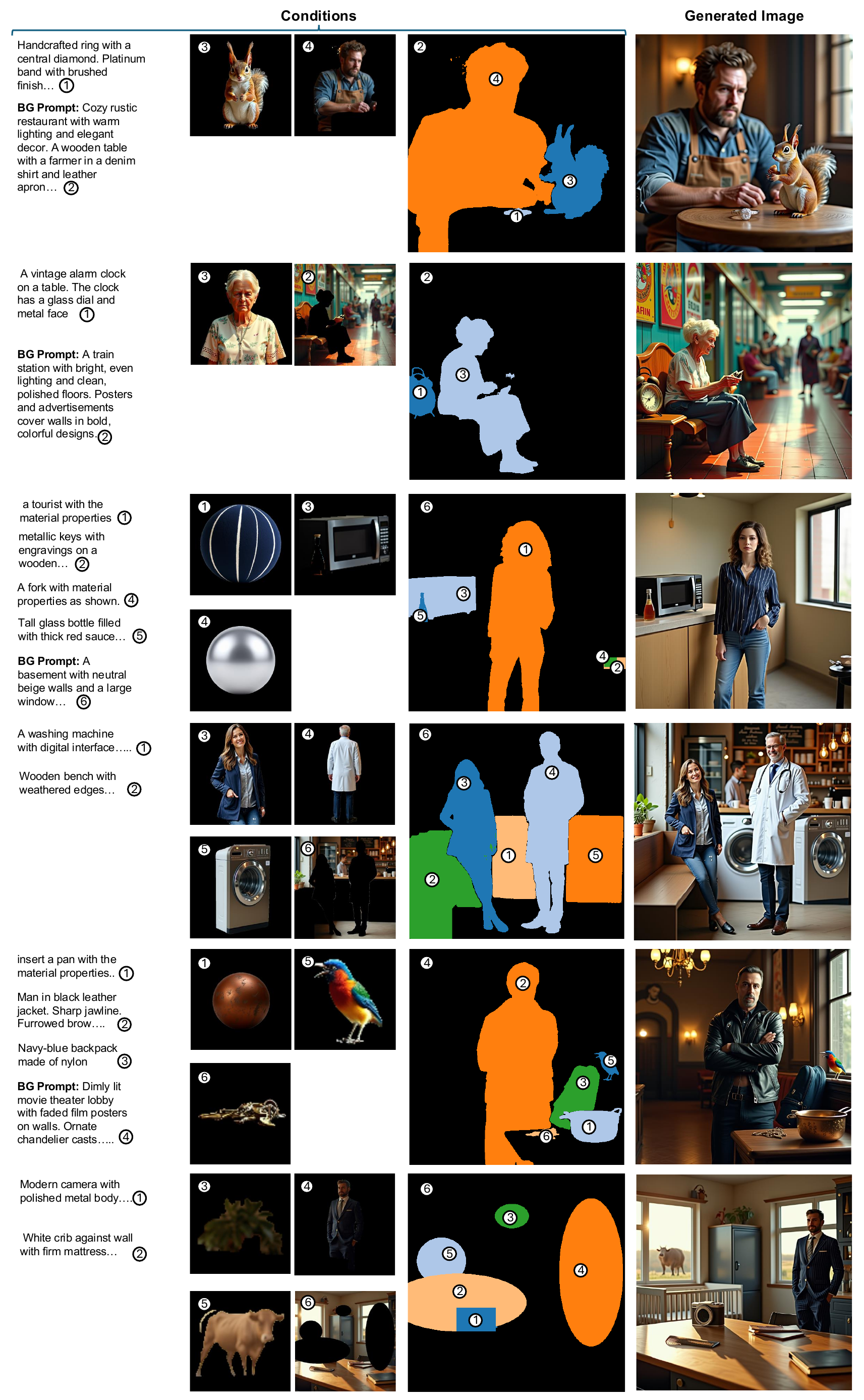}
  \vspace{-0.15in}
  \caption{\textbf{Image Generation and Editing Gallery} 
  \coolname can perform regional edits and generation using heterogeneous conditions in the wild.
  Each region is accompanied by a number which relates the condition to the region.
  }
  \label{fig:qual_gallery_2}
\end{figure*}

%% file: sec/figures_and_tables_tex/fig_supp_gallery.tex
\begin{figure*}[t]
  \centering
  \vspace{-0.15in}
  \includegraphics[width=0.95\linewidth]{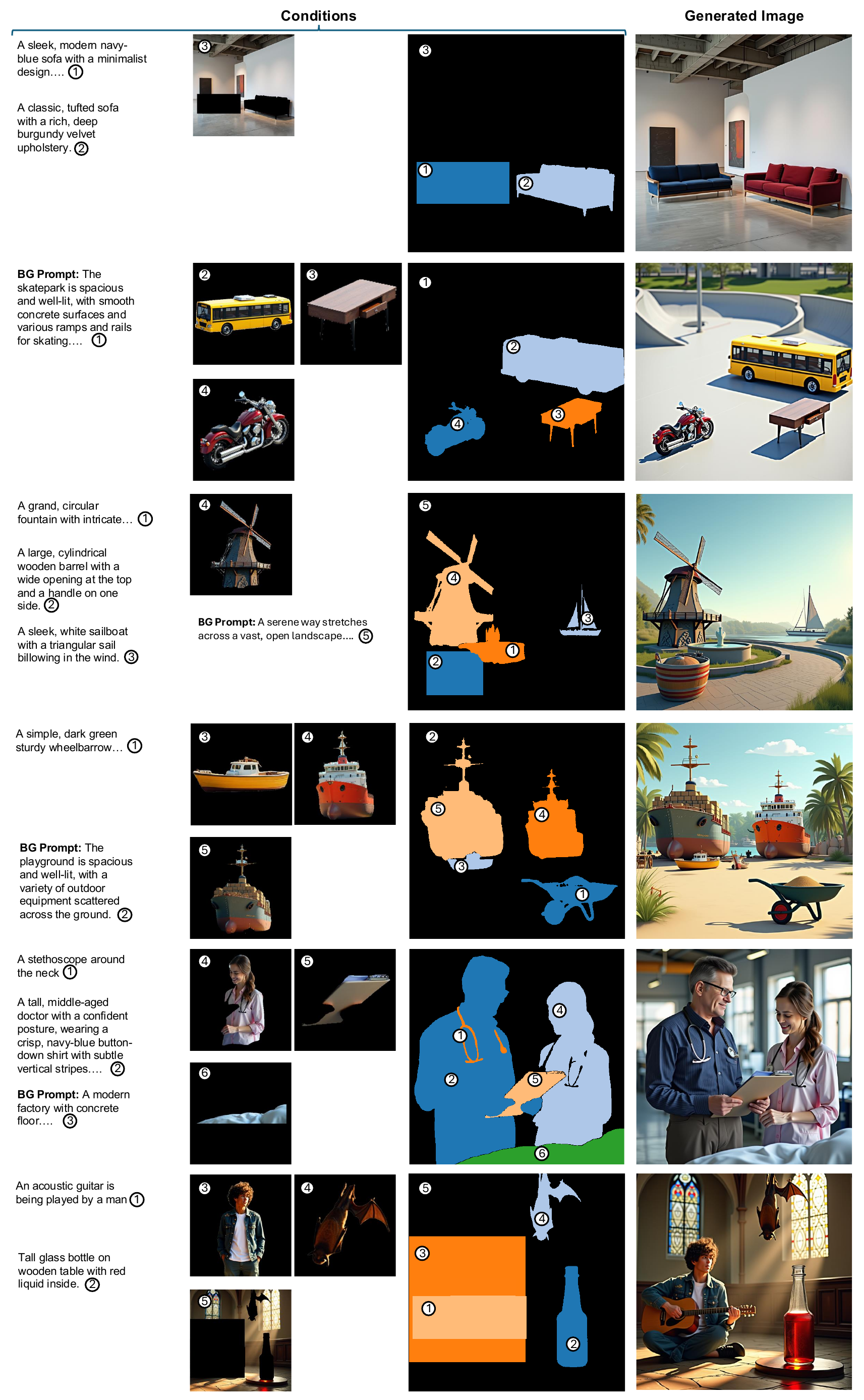}
  \vspace{-0.15in}
  \caption{\textbf{Image Generation and Editing Gallery} 
  \coolname can perform regional edits and generation using heterogeneous conditions in the wild.
  Each region is accompanied by a number which relates the condition to the region.
  }
  \label{fig:qual_gallery_1}
\end{figure*}

%% file: sec/figures_and_tables_tex/fig_supp_gallery_3.tex
\begin{figure*}[t]
  \centering
  \vspace{-0.15in}
  \includegraphics[width=0.95\linewidth]{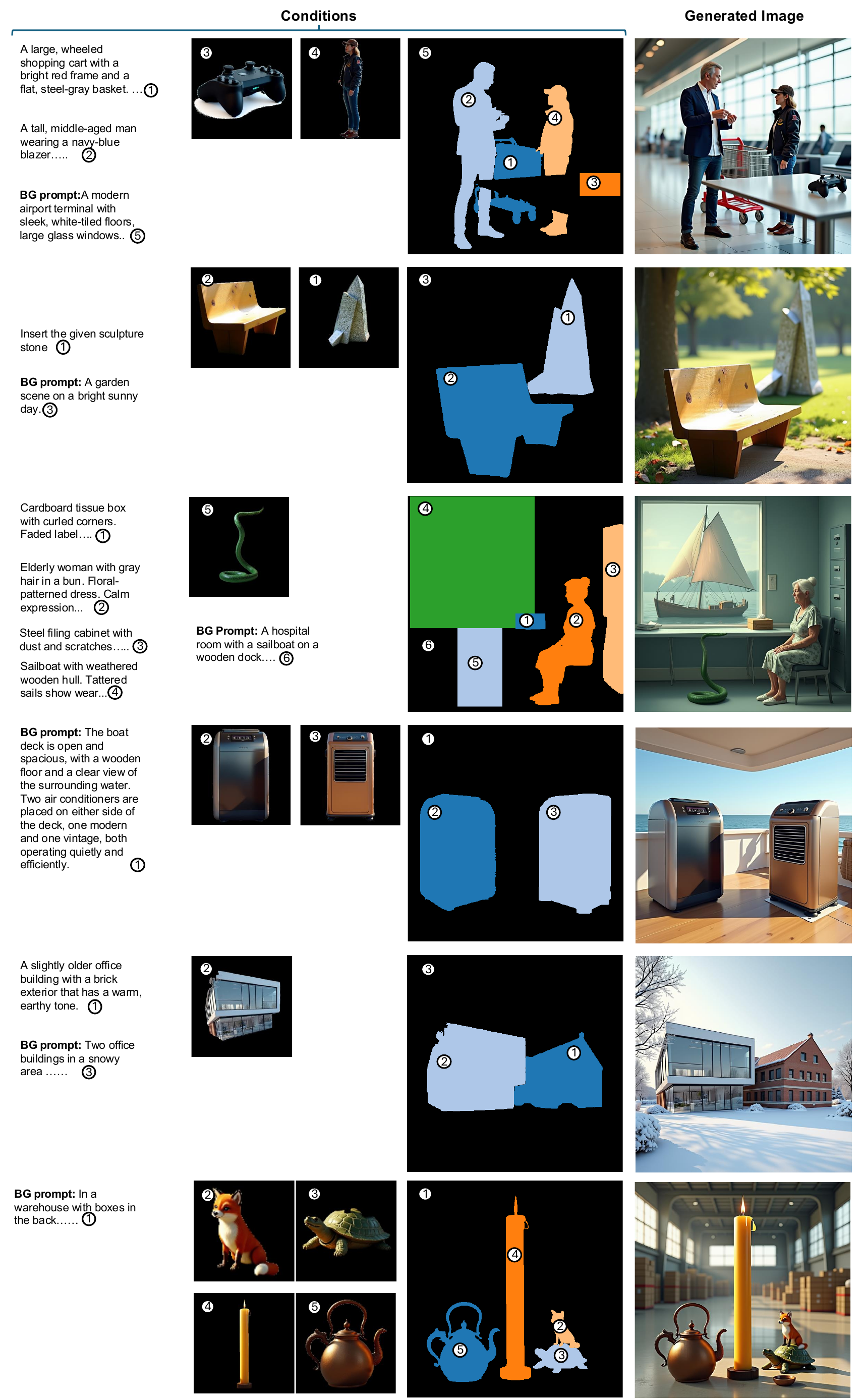}
  \vspace{-0.15in}
  \caption{\textbf{Image Generation and Editing Gallery} 
  \coolname can perform regional edits and generation using heterogeneous conditions in the wild.
  Each region is accompanied by a number which relates the condition to the region.
  }
  \label{fig:qual_gallery_3}
\end{figure*}

%% file: sec/figures_and_tables_tex/fig_sup_in_the_wild_edit.tex
\begin{figure*}[t]
  \centering
  \includegraphics[width=\linewidth]{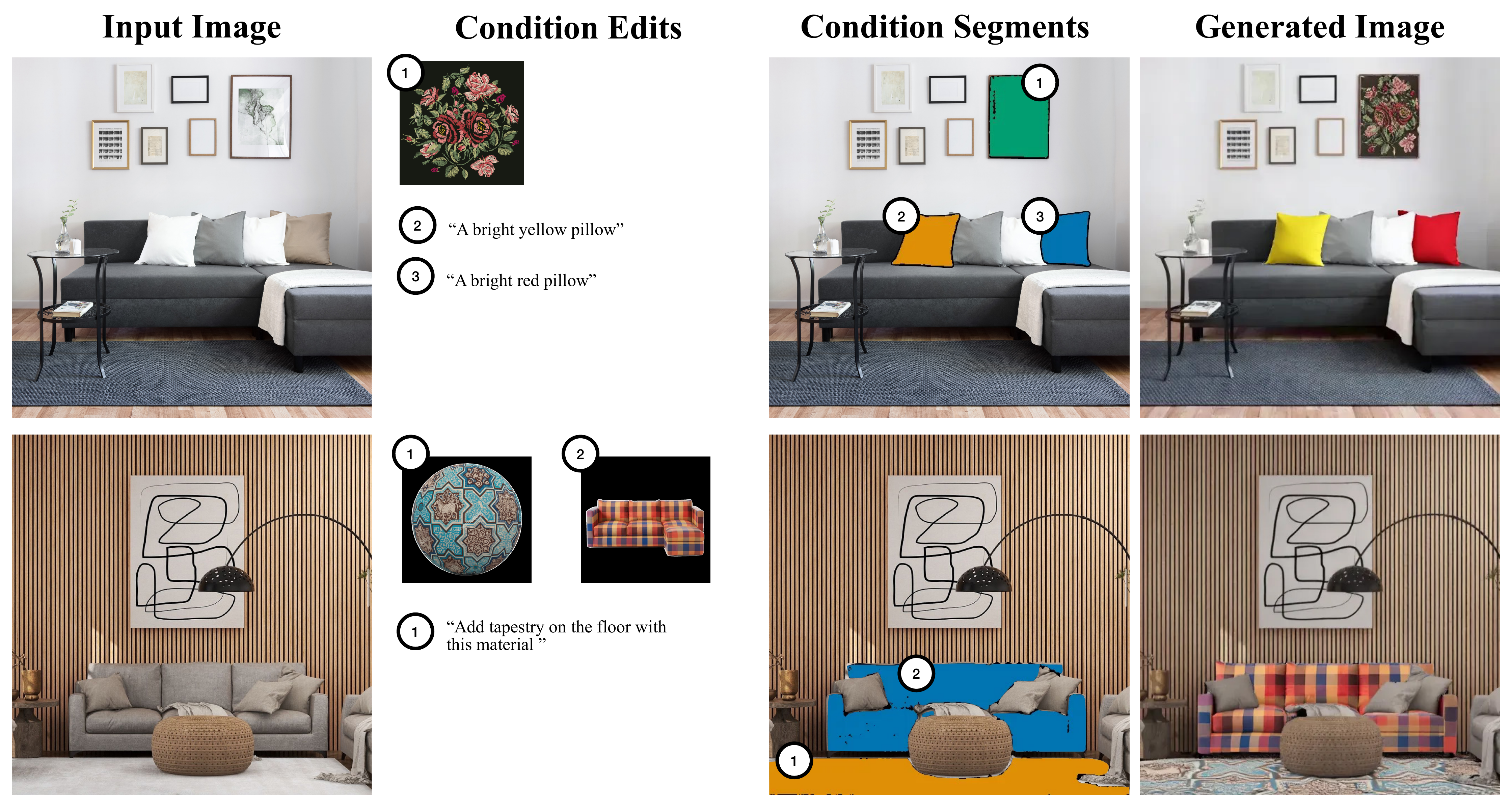}
  \caption{\textbf{Image Editing with Images and Region Prompts.} 
  \coolname can perform one-shot regional edits on images from the internet.
  }
  \label{fig:qualitative_gen_in_the_wild_edit}
\end{figure*}

%% file: sec/figures_and_tables_tex/fig_sup_layout_generation.tex
\begin{figure*}[t]
  \centering
  \includegraphics[width=\linewidth]{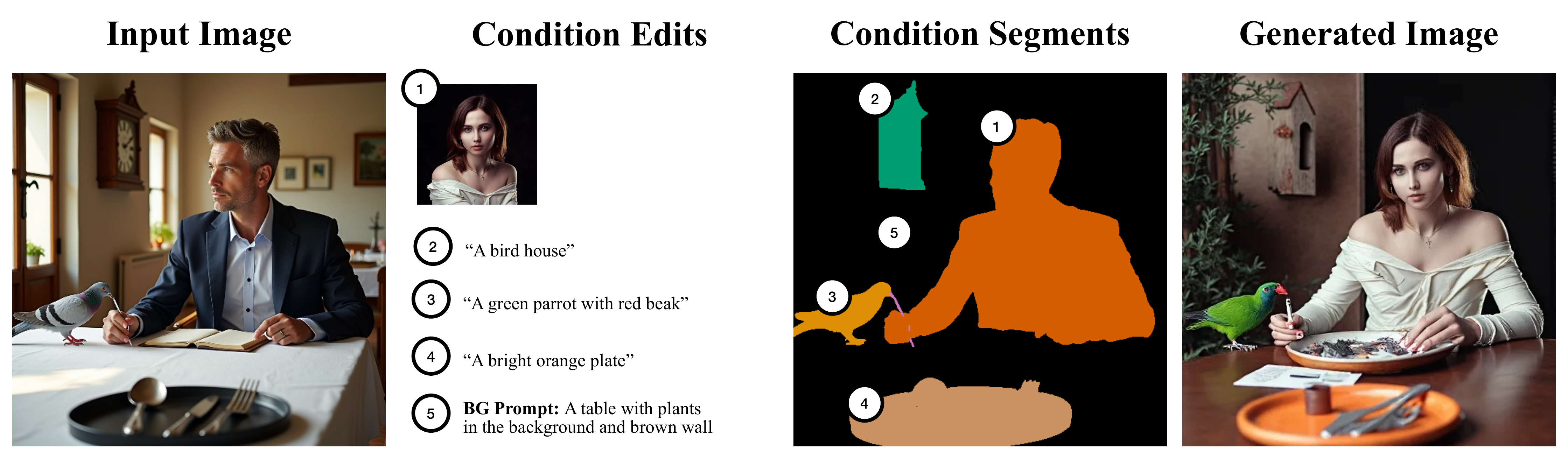}
  \caption{\textbf{Layout Guided Generation.} \coolname can perform layout guided image generation. In this example, an image segmentation model is used to identify region of an input image. Then, we select a subset of these regions and generate a new image following the same layout.
  Region 1 is described by a new subject image, while the other regions have text descriptions.
  }
  \label{fig:qualitative_gen_layout_gen}
\end{figure*}

%% file: sec/figures_and_tables_tex/table_new_distribution.tex
\begin{table}[h]
\centering
\vspace{-0.15in}
\caption{Quantitative comparison on SACap-eval dataset (real-world images). 
Our work is competitive even when trained entirely on synthetic data with better aesthetics and image-text adherance (Zero-Shot setting). 
Minimal fine-tuning with only \textbf{2.3\%} of real-data improves region text adherence significantly.
AppearancePointer is the only model that supports both image and text as regional guidance with the same weights (highlighted in \textcolor{red}{red}).
Best results in gold, second best in silver. 
\label{tab:new_distribution}
}
\vspace{-0.15in}
\resizebox{\linewidth}{!}
{
\begin{tabular}{lccccc|cc}
\toprule
 &
\textbf{Training}
 &
\multicolumn{4}{c}{\textbf{Region}} &
\multicolumn{2}{c}{\textbf{Global}} \\
&
\textbf{Data}
& 
\textbf{CLIP-I$\uparrow$} &
\textbf{CLIP-T$\uparrow$} & \textbf{DINO-I$\uparrow$} & \textbf{MIoU$\uparrow$} &
\textbf{CLIP-T$\uparrow$} & \textbf{CLIP-IQA$\uparrow$} \\
\midrule
\textbf{Task: Region+Text to Image}  \\
\midrule
Seg2Any & SACap-1M & $\times$
& \cellcolor{goldcell} 26.24 & $\times$ 
& 
\cellcolor{goldcell} 64.24 
&\cellcolor{silvercell} 32.30 & 86.07 \\
\textcolor{red}{Ours (Zero-Shot)} & AP-37K &  
$\times$ 
& 25.57 & 
$\times$ 
& 60.14  & 32.17 & \cellcolor{goldcell}89.27  \\
Ours (2.3\% Fine-Tune) & AP-37K + SACap-23K & 
$\times$ 
&\cellcolor{silvercell} 25.78 & 
$\times$ 
& \cellcolor{silvercell}61.72  & \cellcolor{goldcell} 32.99 & \cellcolor{silvercell}87.92 \\
\midrule
\textbf{Task: Region+Image to Image}  \\
\midrule
MS-Diffusion (Bbox) & Private-3M &
87.81 &
$\times$ & 
46.86 &
53.53 &
31.95 &
74.91 \\
\textcolor{red}{Ours (Zero-Shot, Bbox)} & AP-37K & \cellcolor{silvercell} 93.92 & 
$\times$ 
& \cellcolor{silvercell} 72.16 & \cellcolor{silvercell} 59.64 & \cellcolor{silvercell} 32.78 & \cellcolor{silvercell} 86.23 \\
\textcolor{red}{Ours (Zero-Shot, Fine-Grained)} & AP-37K &\cellcolor{goldcell} 94.09 & 
$\times$ 
& \cellcolor{goldcell} 72.63 & \cellcolor{goldcell} 62.46 & \cellcolor{goldcell} 33.02 & \cellcolor{goldcell} 86.27 \\

\bottomrule
\end{tabular}
}
\vspace{-0.2in}
\end{table}

%% file: sec/figures_and_tables_tex/fig_rebuttal_eccv.tex
\begin{figure}[h]
  \centering
  \includegraphics[width=\linewidth]{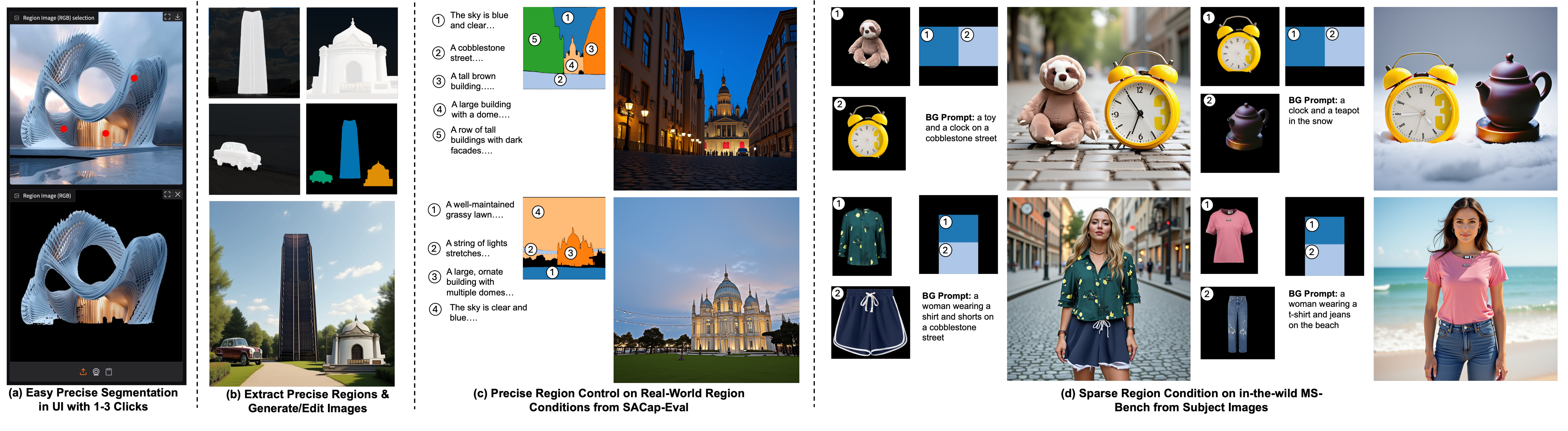}
  \vspace{-0.35in}
  \caption{\textbf{(a)} Precise Easy Segmentation in UI (clicks visualized in red). 
  \textbf{(b)} Obtaining precise regions from other images using a single click and using them as conditions for generating/editing images.
  \textbf{(c)} Real-world regions+text to image generation zero-shot with AppearancePointers on SACap-eval. 
  \textbf{(d)} Sparse Bbox subject conditioned image generation from MS-Bench.
  }
  \vspace{-0.3in}
  \label{fig:rebuttal_eccv}
\end{figure}

%% file: sec/figures_and_tables_tex/fig_sup_failure.tex
\begin{figure}[t]
  \centering
  \includegraphics[width=\linewidth]{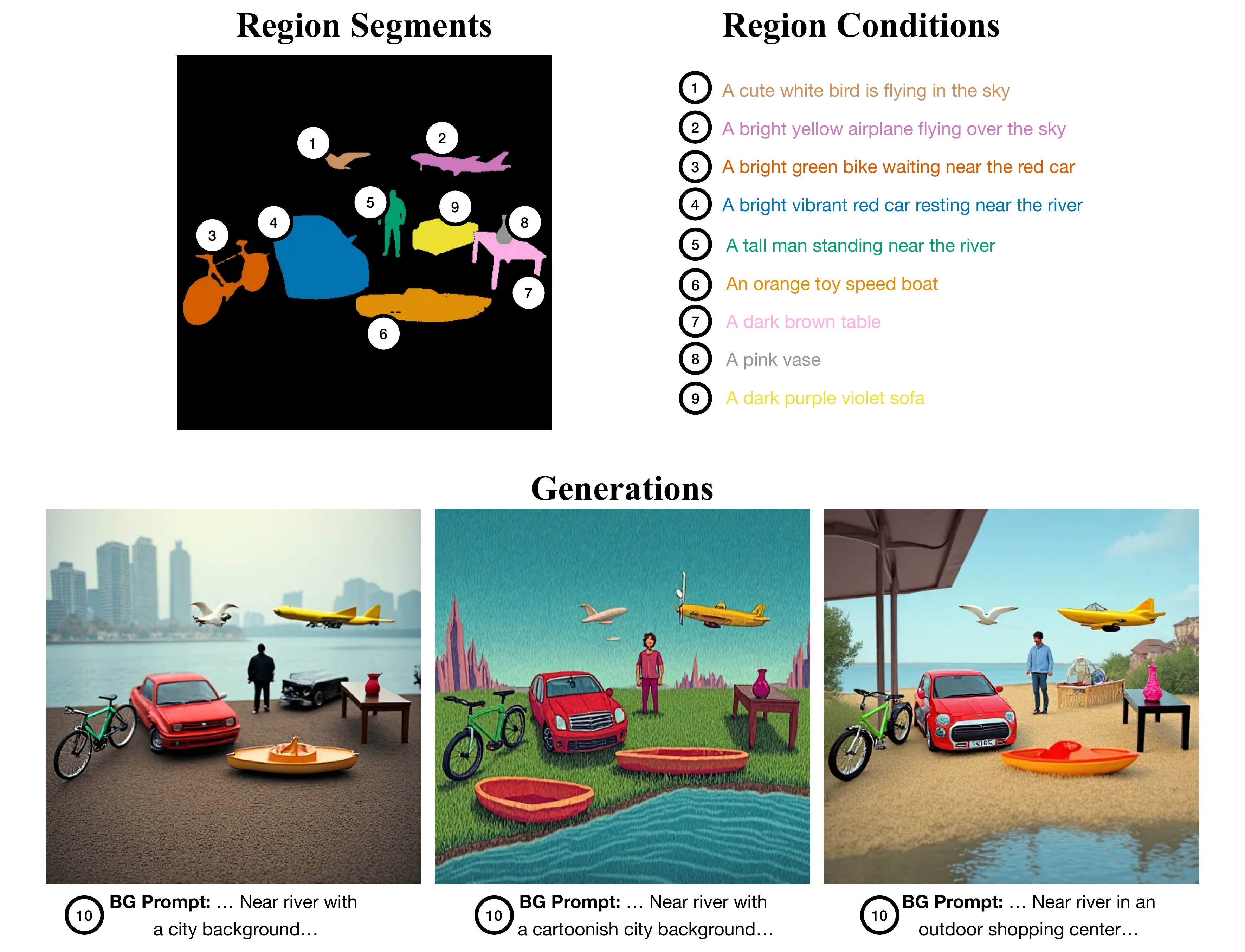}
  \caption{\textbf{Failure Cases.} 
  We test scaling the number of regions using text prompt for \coolname. 
  Increasing number of regions upto 10 in this case can degrade region adherence and quality. 
  Here, the sofa is not captured accurately. 
  }
  \label{fig:qualitative_failure}
\end{figure}

%% file: sec/figures_and_tables_tex/fig_attention_correspondence_supp.tex
\begin{wrapfigure}{r}{0.5\textwidth}  
  \centering
  \vspace{-0.65in}
  \includegraphics[width=\linewidth]{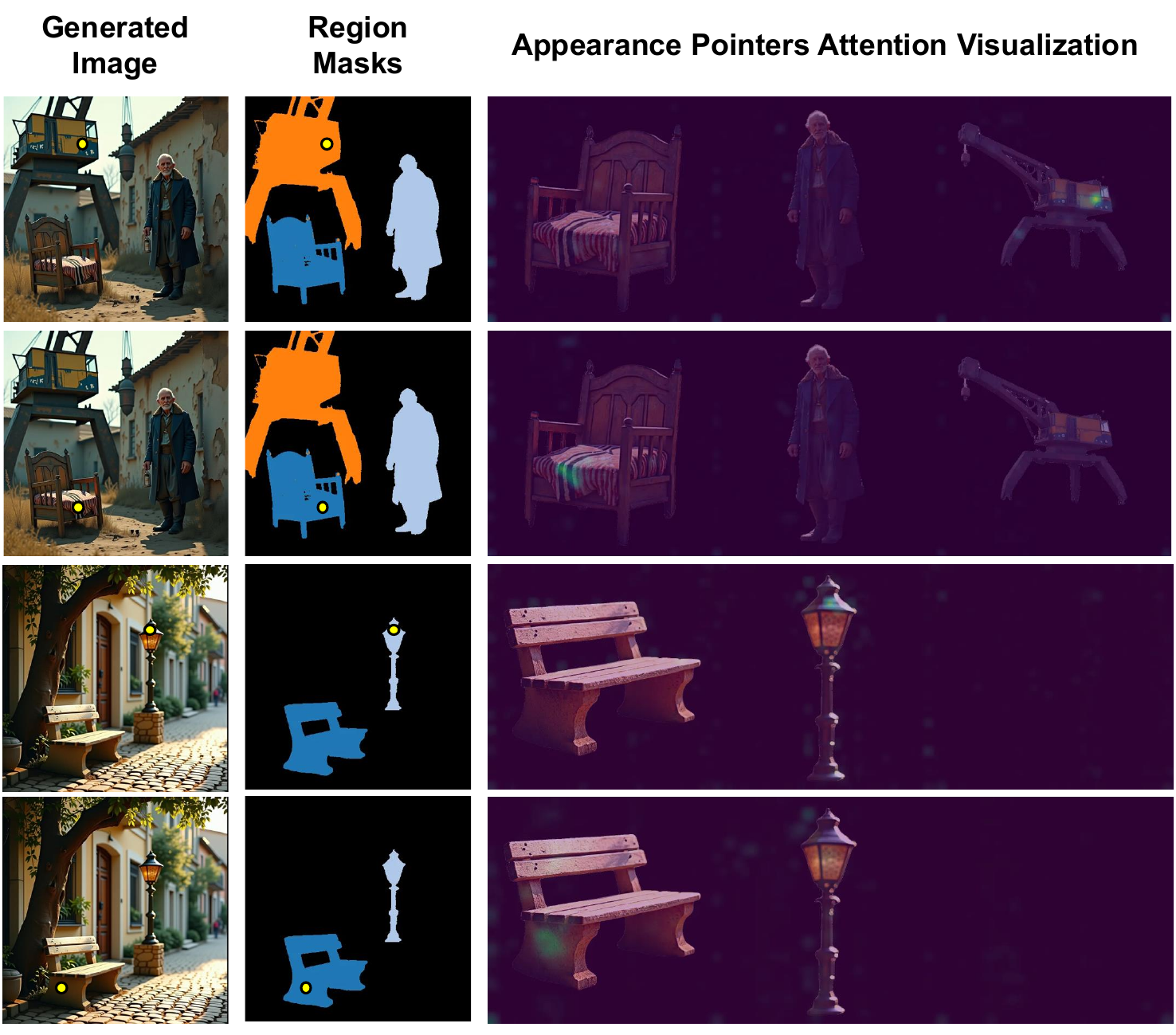}
  \vspace{-0.3in}
  \caption{Appearance Pointer \textit{Points} to Correct Regions of Interest displaying its efficacy in multimodal region-controlled generation.}
  \vspace{-0.3in}
  \label{fig:appearance_pointer_attention_correspondence_supp}
\end{wrapfigure}

%% file: sec/figures_and_tables_tex/novel_view.tex
\begin{listing}
\vspace{0.2in}
\caption{Example Novel Pose Verification JSON \label{json:novel_pose}}
\begin{lstlisting}[language=json]
{
    "overall_consistency_score": 0.8,
    "is_geometrically_consistent": true,
    "sharpness_comparison_score": 0,
    "is_novel_pose": false,
    "overall_consistency_reason": "Good Match (Clear resemblance, but minor, subtle inconsistencies in material or lighting are noticeable).",
    "geometric_consistency_reason": "N/A",
    "sharpness_comparison_reason": "N/A",
    "pose_novelty_reason": "N/A"
}
\end{lstlisting}
\end{listing}

%% file: sec/figures_and_tables_tex/material.tex
\begin{listing}
\caption{Example Material Verification JSON \label{json:material}}
\vspace{-0.05in}
\begin{lstlisting}[language=json]
{
    "perceptual_similarity_score": 0.7,
    "albedo_color_fidelity_score": 0.8,
    "roughness_gloss_fidelity_score": 0.6,
    "texture_detail_fidelity_score": 0.8,
    "albedo_color_reason": "Sphere's color is slightly lighter than the object.",
    "roughness_gloss_reason": "Sphere appears smoother than the object's material.",
    "texture_detail_reason": "Sphere's pattern is similar but lacks ....",
    "perceptual_consistency_reason": "Good match but some differences in texture depth and glossiness."
}
\end{lstlisting}
\end{listing}



%% file: sec/figures_and_tables_tex/prompt_generation.tex
\begin{listing}
\centering 
\caption{Example for Generating Prompts and Object Descriptions for our Dataset \label{json:prompt_generation}}
\vspace{-0.1in}
\begin{lstlisting}[language=json]
{
    "task": "expert scene captioning and material extraction for 3D generation",
    "style": "{style}",
    "environment": "{environments}",
    "required_object": "{prompt_obj_string}",
    "people_details": "{people_detail_description}",
    "constraints": {
        "output_format": "JSON object",
        "total_word_count_min": 1000,
        "total_word_count_max": 3000,
        "scene_caption_word_count": 300,
        "image_quality": "avoid dark, low-resolution, or overly complex scenes",
        "visual_distinctiveness_rule": "ALL objects generated in the scene must be visually differentiable. Ensure colors, primary textures, and unique features DO NOT overlap between objects. For example, if Object A is 'smooth, dark mahogany,' Object B must not be 'smooth, dark walnut.'",
        "material_specificity_rule": "Material descriptions must be vivid, focusing on sensory qualities: the degree of shine, the feel of the texture, and how light interacts with the surface. Make sure you mention description of the dominant color and any texture elements. Avoid technical 3D terms."
    },
    "instructions": [{
        "step": 1,
        "name": "detailed_object_captions",
        "goal": "Generate a detailed, purely visual description (max 100 words per object) for each object listed in 'required_objects'. These descriptions must be entirely isolated: contain NO environmental context, NO relation to neighboring objects, and NO scene placement. Focus only on intrinsic visual attributes.",
        "subtasks": ["For the 'caption' field, detail color, shape, and intricate surface patterns.", 
                     "For the 'dino_caption' field, generate a succinct, precise phrase (under 10 words) focusing on the object's function or unique geometry to maximize detection accuracy, avoiding reliance on simple color.", 
                     "**If the object is a person, the 'material_caption' must describe the material and texture of their most prominent article of clothing (e.g., shirt, jacket, or pants), detailing its weave, reflectivity, and drape.**. For all other objects, provide a rich, sensory and perceptual description of the object's material properties, describing texture, reflectivity, and light interaction in the material caption with the **single** most prominent article of the material along with its color (under 10 words). DO NOT describe multiple materials or colors."
        ]},
        {
        "step": 2,
        "name": "scene_caption",
        "goal": "Generate a final, vivid scene caption by seamlessly combining all objects from Step 1. Focus on **scene-level details** (placement, lighting, composition, object interaction) within the specified **environment** and **style**. Ensure all objects are in sharp focus and not part of a blurred background. The caption must be upto **300 words** long."
        }
    ],
    "output_structure": {
        "detailed_object_captions": {
        "object_1":
            {
                "name": "<object_1_name>",
                "caption": "<detailed object 1 visual caption>",
                "dino_caption": "<succinct dino caption for object 1 for better detection>",
                "material_caption": "<detailed visual and textural description of the object's material>"
            },
        "object_2":
            {
                "name": "<object_2_name>",
                "caption": "<detailed object 2 visual caption>",
                "dino_caption": "<succinct dino caption for object 2 for better detection>",
                "material_caption": "<detailed visual and textural description of the object's material>"

            },
        },
        "scene_caption": "<scene caption, upto 300 words long>",
        "scene_background_caption": "<scene background caption, upto 300 words long>"
    }
}
\end{lstlisting}

\end{listing}

%% file: sec/figures_and_tables_tex/novel_view_rating.tex
\begin{listing}
\caption{Example for Rating the Quality of Novel View Pose \label{json:novel_view_rating}}
\vspace{-0.05in}
\begin{lstlisting}[language=json]
{
    "task": "Expert simplified 3D view consistency and critical artifact analysis",
    "goal": "Given two images of the same object from different viewpoints, provide a single overall consistency score and three critical binary/categorical checks (Geometric, Sharpness, Pose). The ONLY output must be a single, valid JSON object that adheres strictly to the defined output_structure, with absolutely no preamble, explanation, or concluding text.",
    "prompt": "Analyze the two object views. Assume both images are renders of the *same 3D model* and *same material*. **CRITICAL RULE: IGNORE ALL PIXELATION AND BLOCKING ARTIFACTS** when determining scores. On a scale from 0.0 (inconsistent) to 1.0 (perfectly consistent), score the **overall visual consistency**, encompassing perceptual match, material realism, and texture detail. **GEOMETRIC CHECK:** Set **is_geometrically_consistent** to **true** if the 3D shape/scale is plausible and consistent, and **false** if there are significant errors. Provide an **integer score** for the sharpness comparison based on the rule: **0** if SAME SHARPNESS; **-1** if View 1 is MORE BLURRY; and **+1** if View 2 is MORE BLURRY. Additionally, set **is_novel_pose** to **true** if View 1 and View 2 have a DIFFERENT POSE or ORIENTATION, and **false** otherwise. Provide a brief, one-sentence justification for **every output**. **MAKE SURE YOU STICK TO THE RUBRIK AND FOLLOW IT EXACTLY**. **IGNORE THE WHITE BACKGROUND IN BOTH IMAGES. JUST FOCUS ON THE OBJECT ITSELF**.",
    "scoring_rubric": {
        "overall_consistency_score": {
            "0.9-1.0": "Excellent Match (Nearly indistinguishable; excellent consistency across all lighting, texture, and perceptual aspects, ignoring pixelation).",
            "0.7-0.89": "Good Match (Clear resemblance, but minor, subtle inconsistencies in material or lighting are noticeable).",
            "0.4-0.69": "Fair Match (Obvious inconsistencies in one or more areas, such as color shift or smeared textures).",
            "0.0-0.39": "Poor Match (Gross inconsistencies; views look like different objects or materials)."
        },
        "is_geometrically_consistent": {
            "true": "3D shape, scale, and placement are plausible and consistent (minor artifacts acceptable).",
            "false": "Significant geometric errors are present (distortion, object shifting, hole-filling errors, or wrong scale)."
        },
        "sharpness_comparison_score": {
            "+1": "View 2 is noticeably softer, has less fine detail, or appears more out-of-focus than View 1 (View 2 is the blurrier image).",
            "0": "The perceptual sharpness and fine detail levels are visually identical between View 1 and View 2.",
            "-1": "View 1 is noticeably softer, has less fine detail, or appears more out-of-focus than View 2 (View 1 is the blurrier image)."
        },
        "is_novel_pose": {
            "true": "**NOVEL POSE**: View 2 is rotated/transformed and is different from view 1 (difference is easily visible).",
            "false": "**NO NOVELTY IN POSE**: View 2 is the same orientation as view 1 (difference is minimal or invisible)."
        }
    },
    "output_structure": {
        "overall_consistency_score": "float (0.0 to 1.0)",
        "is_geometrically_consistent": "bool (one of: true, or false)",
        "sharpness_comparison_score": "integer (one of: -1, 0, or 1)",
        "is_novel_pose": "bool (one of: true, or false)",
        "overall_consistency_reason": "string (e.g., 'Low score due to clear color shift and texture smearing, NOT pixelation' or 'N/A')",
        "geometric_consistency_reason": "string (e.g., 'Significant geometric distortion is visible in View 2' or 'N/A')",
        "sharpness_comparison_reason": "string (e.g., 'Score is +1 because fine edges are lost in View 2' or 'N/A')",
        "pose_novelty_reason": "string (e.g., 'Pose is novel because it is rotated 30 degrees around up-down axis' or 'N/A')"
    }
}
\end{lstlisting}

\end{listing}

%% file: sec/figures_and_tables_tex/material_rating.tex
\begin{listing}
\caption{Example for Rating the Material/Texture Quality of the Generated Edit\label{json:material_rating}}
\vspace{-0.05in}
\begin{lstlisting}[language=json]
{
    "task": "expert graphics 3D shape and material understanding",
    "goal": "numerically score the fidelity of three distinct PBR attributes between the object and sphere images.",
    "prompt": "Analyze the two images (segmented object and sphere). On a scale from 0.0 to 1.0, score the degree to which the material properties and texture on the sphere accurately represent those of the object. Score the **Albedo/Color**, **Roughness/Gloss**, **Perceptual Match** and **Texture/Pattern Detail** independently. Output the three scores as floating-point numbers. **MAKE SURE YOU STICK TO THE RUBRIK AND FOLLOW IT EXACTLY**. **IGNORE THE WHITE BACKGROUND IN BOTH IMAGES. JUST FOCUS ON THE OBJECT AND SPHERE ITSELF**.",
    "scoring_rubric": {
        "perceptual_similarity": {
            "0.9-1.0": "Excellent Match (Visually indistinguishable; the sphere perfectly captures the 'feel' and realism of the object's material).",
            "0.7-0.89": "Good Match (The sphere clearly represents the intended material, but a layman could spot minor flaws like poor lighting interaction or blurring).",
            "0.4-0.69": "Fair Match (The material looks synthetic, plastic, or simply 'off' compared to the real object, severely impacting believability).",
            "0.0-0.39": "Poor Match (The material is unrecognizable or looks like a gross simplification; fails the basic realism test)."
        },
        "albedo_color": {
            "0.9-1.0": "Excellent Match (Nearly identical color, saturation, and lightness).",
            "0.7-0.89": "Good Match (Correct hue, but lightness or saturation is clearly wrong/off).",
            "0.4-0.69": "Fair Match (Significant color shift; related but incorrect hue).",
            "0.0-0.39": "Poor Match (Fundamentally different color)."
        },
        "roughness_gloss": {
            "0.9-1.0": "Excellent Match (Highlights and surface diffusion perfectly match observed shininess/dullness).",
            "0.7-0.89": "Good Match (Generally correct, but highlights are slightly too sharp or too broad).",
            "0.4-0.69": "Fair Match (Significant misrepresentation of surface type, e.g., shiny plastic rendered as dull cloth).",
            "0.0-0.39": "Poor Match (Completely inverted property)."
        },
        "pattern_detail": {
            "0.9-1.0": "Excellent Match (Texture pattern, scale, and orientation are identical).",
            "0.7-0.89": "Good Match (Correct pattern used, but slightly tiled, scaled, or rotated incorrectly).",
            "0.4-0.69": "Fair Match (Right *type* of texture, but the *specific pattern* is different or map resolution is poor).",
            "0.0-0.39": "Poor Match (Completely different pattern applied)."
        }
    },
    "output_structure": {
        "perceptual_similarity_score": "float (0.0 to 1.0)",
        "albedo_color_fidelity_score": "float (0.0 to 1.0)",
        "roughness_gloss_fidelity_score": "float (0.0 to 1.0)",
        "texture_detail_fidelity_score": "float (0.0 to 1.0)",
        "albedo_color_reason": "string (e.g., 'Sphere's color is too light' or 'N/A')",
        "roughness_gloss_reason": "string (e.g., 'Sphere is too shiny' or 'N/A')",
        "texture_detail_reason": "string (e.g., 'Sphere's pattern is different than object as it contains....' or 'N/A')",
        "perceptual_consistency_reason": "string (e.g., 'Low perceptual similarity because..... ' or 'N/A')"

    }
}
\end{lstlisting}

\end{listing}